\documentclass{article}

\PassOptionsToPackage{numbers, compress}{natbib}

\usepackage[final]{neurips_2021}

\usepackage[utf8]{inputenc} 
\usepackage[T1]{fontenc}    
\usepackage{hyperref}       
\usepackage{url}            
\usepackage{booktabs}       
\usepackage{amsfonts}       
\usepackage{nicefrac}       
\usepackage{microtype}      
\usepackage{xcolor}         
\usepackage{amsmath}
\usepackage{graphicx}
\usepackage{algorithm}
\usepackage{algorithmic}
\usepackage{multirow} 
\usepackage{enumerate}
\usepackage{setspace}
\usepackage{booktabs}
\usepackage{subcaption}
\usepackage{diagbox}
\usepackage{wrapfig,lipsum}
\usepackage{amsmath,amsthm,amsfonts,amssymb,mathtools}
\usepackage{authblk}
\newtheorem{theorem}{Theorem}[section]

\title{R-Block: Regularized block of Dropout for convolutional networks}

\author{%
	\textbf{Liqi Wang}\textsuperscript{\normalfont 1} \quad
	\textbf{Qiya Hu}\textsuperscript{\normalfont 1}\thanks{Corresponding author.}
    
    \textsuperscript{1}Academy of Mathematics and Systems Science, Chinese Academy of Sciences
}

\begin{document}
\renewcommand{\thefootnote}{\fnsymbol{footnote}}
\setcounter{footnote}{-1}
\footnote{ \textbf{Funding}: The work of the author was supported by the National Natural Science Foundation of China grant G12071469. 
}

\maketitle

\begin{abstract}
Dropout as a regularization technique is widely used in fully connected layers while is less effective in convolutional layers. Therefore more structured forms of dropout have been proposed to regularize convolutional networks. The disadvantage of these methods is that the randomness introduced causes inconsistency between training and inference. In this paper, we apply a mutual learning training strategy for convolutional layer regularization, namely R-Block, which forces two outputs of the generated difference maximizing sub models to be consistent with each other. Concretely, R-Block minimizes the losses between the output distributions of two sub models with different drop regions for each sample in the training dataset. We design two approaches to construct such sub models. Our experiments demonstrate that R-Block achieves better performance than other existing structured dropout variants. We also demonstrate that our approaches to construct sub models outperforms others.

\end{abstract}

\section{Introduction}
Convolutional Neural Networks (CNNs) are powerful tools in the field of computer vision. Regularization~\cite{Ba2016LayerN,Hinton2015DistillingTK,Ioffe2015BatchNA,Szegedy2015RethinkingTI,Wu2018GroupN,Zhang2019BeYO,Li2022ASO} is needed during CNN training to prevent overfitting and improve the generalization ability of the model. Among them, dropout ~\cite{Srivastava2014DropoutAS} is a well-known regularization technique, which can perform implicit ensemble and reduce interdependence of neurons (co-adaptation)~\cite{Li2022ASO} by randomly dropping units in hidden layers during training. Although dropout works well in fully connected layers, it is nearly ineffective in convolutional layers where features are spatially correlated~\cite{Ghiasi2018DropBlockAR}. To address this dropout problem, many structured dropout methods have gradually been proposed~\cite{Tompson2014EfficientOL,Park2016AnalysisOT,Wu2015TowardsDT,Khan2017RegularizationOD,Cai2019EffectiveAE,Hou2019WeightedCD,Zeng2021CorrelationbasedSD,Lu2021LocalDropAH,Pham2021AutoDropoutLD,Chen2020DropClusterAS}. A drawback of these methods is that the randomly sampled sub models during training are different from the full model during inference~\cite{Ma2016DropoutWE,Zolna2017FraternalD}.

In this paper, we adopt a simple and effective strategy to regularize above inconsistency in CNNs, named as R-Block. Specifically, in each mini-batch training, the same sample goes through two sub models with completely different random drop regions and obtains the similar prediction outputs. R-Block minimizes the bidirectional losses between the two output distributions, that is, it adds deep mutual learning~\cite{zhang2018deep} of the two sub models with difference maximization, which can reduce the inconsistency between training and inference phases of model~\cite{Liang2021RDropRD}. According to the structure of feature maps, we design two approaches to construct sub models with different drop regions.

From the perspective of convolutional neural network regularization, our proposed R-Block can be regarded as a new structured form of dropout. In our experiments, R-Block is much better than other existing structured dropout variants. Our results show that our approachs to construct sub models perform better than the alternatives.

\noindent{\bf Related Work.} For CNNs with special structures, structured variants of dropout have been proposed such as
SpatialDropout~\cite{Tompson2014EfficientOL,Park2016AnalysisOT}, Max-pooling Dropout~\cite{Wu2015TowardsDT}, DropBlock~\cite{Ghiasi2018DropBlockAR}, Spectral Dropout~\cite{Khan2017RegularizationOD}, Drop-Conv2d~\cite{Cai2019EffectiveAE}, Weighted Channel Dropout~\cite{Hou2019WeightedCD}, DropCluster~\cite{Chen2020DropClusterAS}, CorrDrop~\cite{Zeng2021CorrelationbasedSD}, LocalDrop~\cite{Lu2021LocalDropAH}, AutoDropout~\cite{Pham2021AutoDropoutLD}, etc. Our method, R-Block, is closely related to SpatialDropout and DropBlock. SpatialDropout randomly masks out entire channels of a feature map and DropBlock randomly drops square regions of a feature map. Since selecting contiguous blocks exploits the spatial correlation in feature maps, these two methods both have good regularization effects in CNNs. We design BDropDML and SDropDML by splitting channels and regions respectively to construct sub models with different dropping units. R-Block enforces the sub models to be consistent with each other to alleviate the training inconsistency. It is inspired by R-Drop~\cite{Liang2021RDropRD} and generalizes the idea of "dropout twice"~\cite{Gao2021SimCSESC} from dropout to structured dropout variants. Our experiments show that R-Block is more effective than R-Drop in CNNs.


\section{R-Block Regularization}
\label{sec:R-Block regularization}

This section introduces our proposed regularization method, R-Block, in detail.
Given the training dataset $\mathcal{D} = \{(x, y)\}$, where $x\in\mathcal{R}^{m\times n\times c} $ and $y\in \{1,2,3,...,\kappa\}$ correspond to the image and its label in the dataset for the image classification task, $m\times n\times c$ is the height $\times$ width $\times$ channel dimension of a feature tensor and $\kappa$ is the number of distinctive classes. The goal of the training is to learn a model $F_\Theta$, where $\Theta$ represents all parameters in a CNN.

Given the input data $(x,y)$ at each training step, we feed $x$ to the forward pass of two sub models $F_\Theta^1$ and $F_\Theta^2$ with different drop regions. Therefore, we can obtain two distributions $p_1$  and $p_2$ of the model predictions, expressed as:
\begin{equation}
	\label{eqn:predictions}
	p^k_i(x)=\dfrac{exp([F^i_\Theta(x)]_k/T)}{\sum_jexp([F^i_\Theta(x)]_j/T)},
\end{equation}
where $p_i^k$  represents the probability of the $k$th class for the $i(i=1,2)$ sub model $F_\Theta^i$, and $T>0$ is the temperature scaling parameter.

The conventional supervised loss between the predicted values of the sub model $F_\Theta^i$  and the correct labels $y$ is the standard cross-entropy loss $J_{ce}$:
\begin{equation}
	\label{eqn:cross-entropy loss}
	J_{ce}(F_\Theta^i(x),y)=-\ln\dfrac{exp([F^i_\Theta(x)]_y)}{\sum_j exp([F^i_\Theta(x)]_j)}.
\end{equation}
We use the Kullback Leibler (KL) Divergence to quantify the match of the output distributions $p_1$  and $p_2$ of two sub models. The KL distance $J_{KL}$  from $p_i$  and $p_j$ is computed as:
\begin{equation}
	\label{eqn:KL loss}
	J_{KL}(p_j,p_i)=\sum_k[p_j]_k\ln\dfrac{[p_j]_k}{[p_i]_k}.
\end{equation}
The overall loss functions $J_i$  for the sub model $F_\Theta^i$ is:
\begin{equation}
	\label{eqn:sub model loss}
	J_i(F_\Theta^i(x),y)=(1-\alpha)J_{ce}(F_\Theta^i(x),y)+\alpha T^2\sum_{j\neq i}J_{KL}(p_j,p_i),
\end{equation}
where $\alpha>0$ is the coefficient weight. Note that we multiply the square of the temperature $T^2$ by following the original KD~\cite{Hinton2015DistillingTK}.

\begin{figure}
	\centering
	\includegraphics[width=0.75\linewidth]{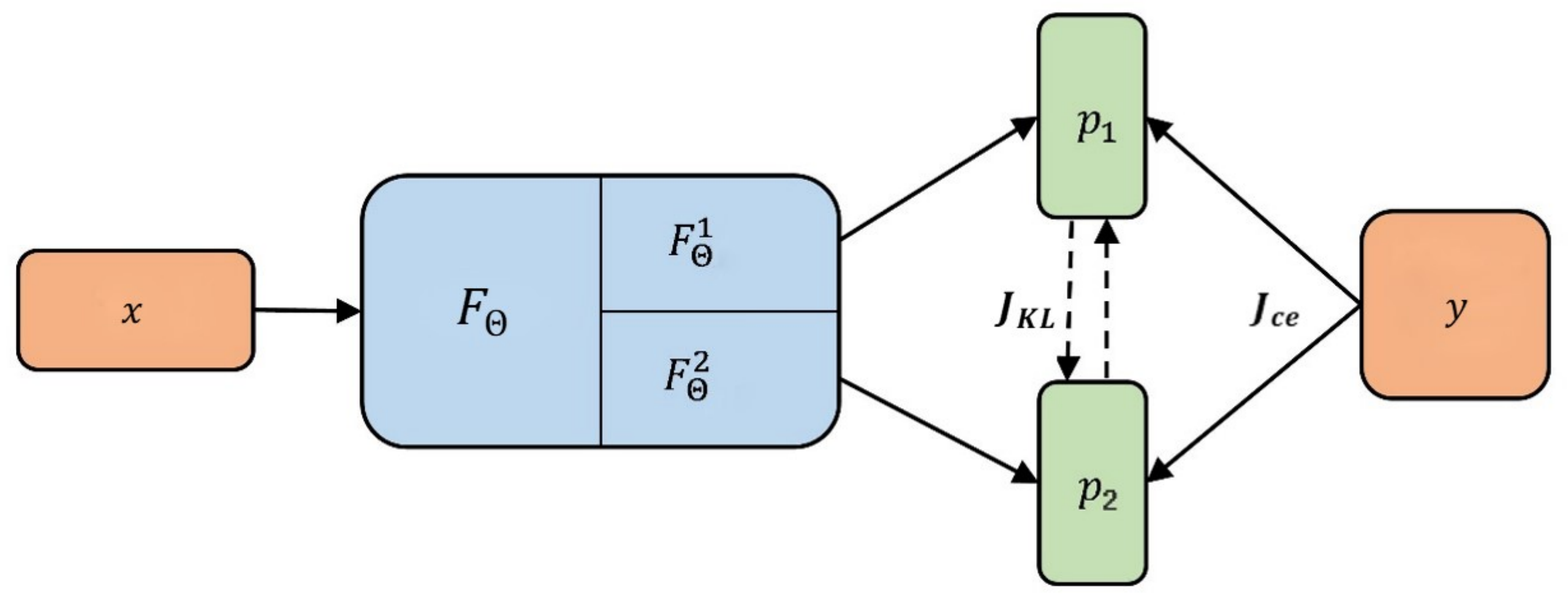}
	\caption{The overall framework of our proposed R-Block. The picture shows that one input $x$ will go through two different sub models and obtain two distributions $p_1$  and $p_2$. The total training loss includes the conventional supervised loss $J_{ce}$ from correct labels and the KL Divergence $J_{KL}$ between the two distributions of two sub models.}
	\label{fig:framework}
\end{figure}

The total training loss $J_{R-Block}$ is defined as follows:
\begin{equation}
	\label{eqn:total loss}
	J_{R-Block}=\dfrac{1}{|\mathcal{D}|}\sum_{(x,y)\in \mathcal{D}}\sum_iJ_i(F_\Theta^i(x),y),
\end{equation}
The overall framework of R-Block is shown in Figure~\ref{fig:framework}. 

\section{Sub models}
\label{sec:sub models}
In this section, we design two approaches, BDropDML and SDropDML, to construct sub models with completely different drop regions. There are two main parameters: $b_{size}$ and $p$. $b_{size}$ is the size of the block to be dropped, and $p$ is the dropout probability for each activation unit.

\begin{figure}
    \centering
    \includegraphics[width=0.68\linewidth]{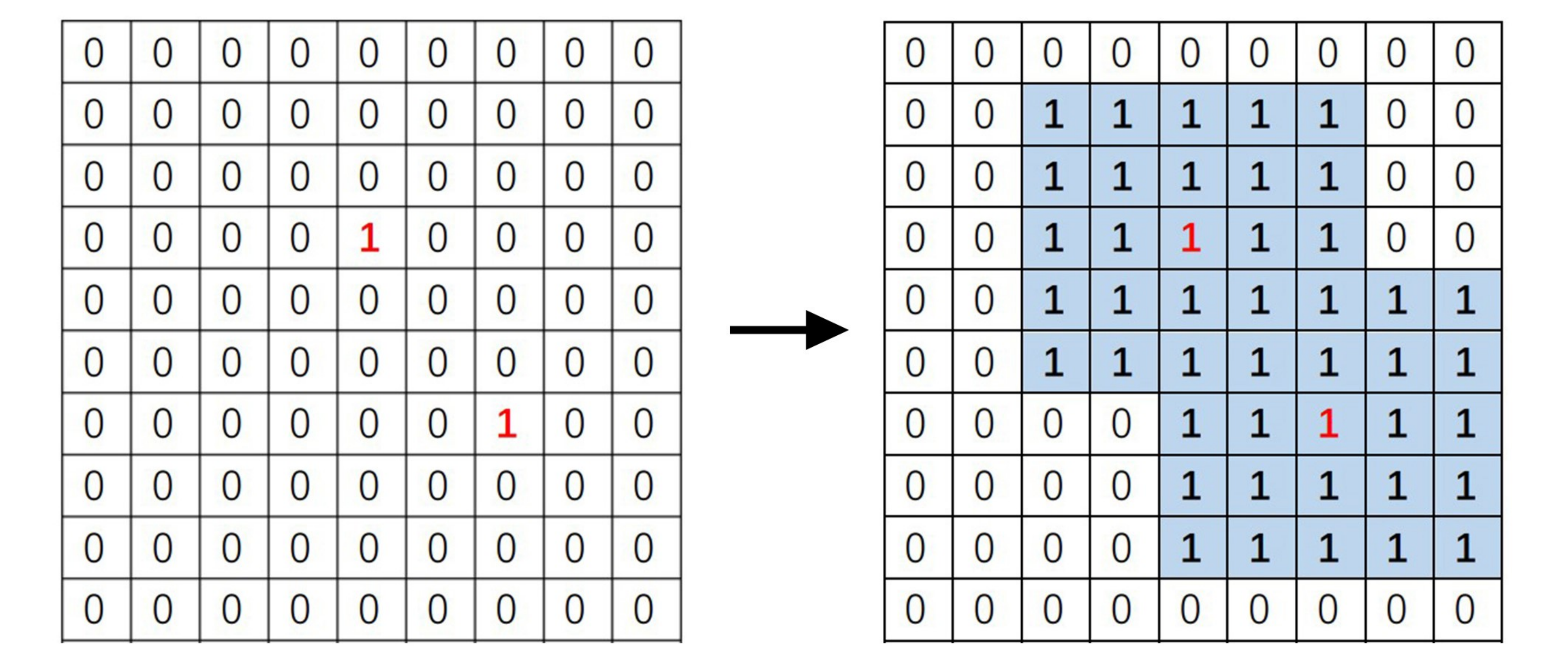}
    \caption{Example of mask sampling, where $b_{size}=5$.}
    \label{fig:block}
\end{figure}

We reference the mask sampling in DropBlock as shown in Figure~\ref{fig:block}: first, we compute the dropout probability $\gamma$ of block center by the dropout probability $p$ of each activation unit, and sample a mask $M$ with probability $\gamma$ on each feature map; second, every $1$ entry on $M$ is expanded to $1$ block of $b_{size}\times b_{size}$. In order to position and maintain symmetry, $b_{size}$ is usually set to an odd number. Here we denote $b_{size}=2k+1$.

\subsection{BDropDML}
\label{sec:BDropDML}
\begin{figure}
	\centering
	\includegraphics[width=0.95\linewidth]{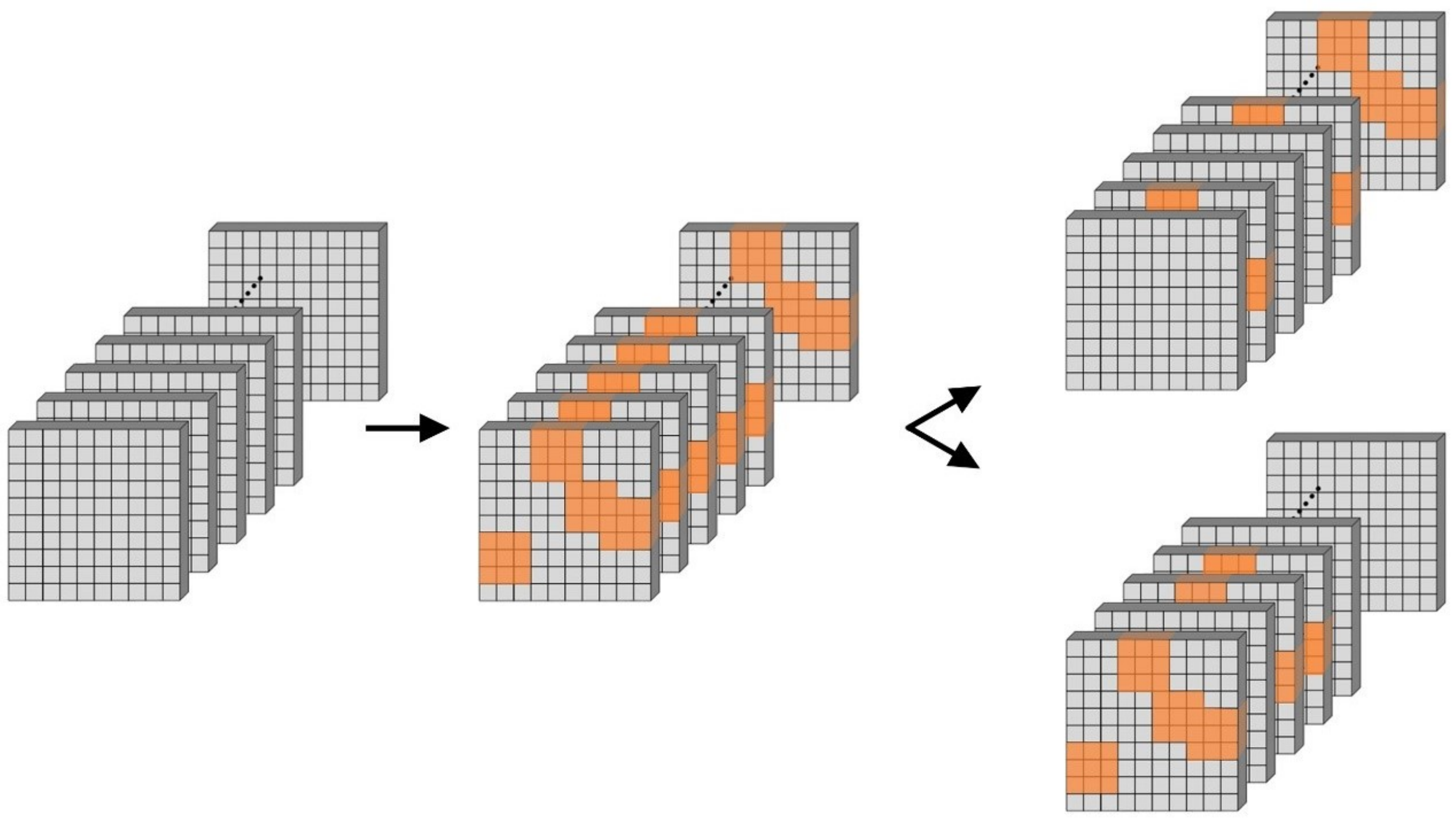}
	\caption{The overall framework of BDropDML.}
	\label{fig:BDropDML}
\end{figure}

We propose a strategy for constructing two sub models with complementary drop regions on feature channels, coined Block Dropout Deep Mutual Learning (BDropDML). Specifically, two sub models share the same DropBlock mask on each feature channel and then perform complementary mask division on channels with a probability of $0.5$. The overall framework of BDropDML is shown in Figure~\ref{fig:BDropDML}, where the yellow Blocks represent drop regions.

Given one input $x\in\mathcal{R}^{m\times n\times c}$, the size $b_{size}$ of the block to be dropped and the dropout probability $p$ for each activation unit, the dropout probability $\gamma$ of block center can be computed as:
\begin{equation}
	\label{eqn:gamma1}
	\gamma = \frac{1}{b_{size}^2}p,
\end{equation}
or:
\begin{equation}
	\label{eqn:gamma2}
	\gamma = \frac{mn}{b_{size}^2(m-b_{size}+1)(n-b_{size}+1)}p.
\end{equation}
The details of BDropDML are given in Algorithm~\ref{al:BDropDML}.

\begin{algorithm}[htbp]
	\setstretch{1.28}
	\caption{BDropDML}
	\label{al:BDropDML}
	\textbf{Input}: $x\in\mathcal{R}^{m\times n\times c}$,  $b_{size}$, $p$. \\
	\textbf{Output}: Maskes sampling $M^1$, $M^2$.
	\begin{algorithmic}[1]
		\STATE Calculate $\gamma$ by Eq.~(\ref{eqn:gamma1}) or Eq.~(\ref{eqn:gamma2}), randomly sample mask $r\in \mathcal{R}^{m\times n}$: \\
		$r_{i,j}\sim\textit{Bernoulli}(1-\gamma)$, $\forall i\in \{1,2,\ldots,m\}$, $j\in \{1,2,\ldots,n\}$.
		\STATE  For each $1$ position $r_{i,j}$, create a spatial square mask $M\in\mathcal{R}^{m\times n}$ with the center being $r_{i,j}$, the width, height being $b_{size}$ and set all the values of $r$ in the square to be $1$.
		\STATE Randomly sample mask $l^1\in\mathcal{R}^c$: $l_k^1\sim\textit{Bernoulli}(0.5)$, $\forall k\in \{1,2,\ldots,c\}$.\\
		Calculate $l^2\in\mathcal{R}^c$: $l^2=\textbf{1}_{l^1}-l^1$.
		\STATE Expand $M$ to $\hat{M}\in\mathcal{R}^{m\times n\times c}$: $\hat{M}_{i,j,k}=M_{i,j}$.\\
		Expand $l^1$, $l^2$ to $\hat{l}^1$, $\hat{l}^2\in\mathcal{R}^{m\times n\times c}$: $\hat{l}^1_{i,j,k}=l_k^1$,  $\hat{l}^2_{i,j,k}=l_k^2$,\\
		$\forall i\in \{1,2,\ldots,m\}$, $j\in \{1,2,\ldots,n\}$, $k\in \{1,2,\ldots,c\}$.
		\STATE Calculate: $\bar{M}^i=\textbf{1}_x-\hat{M}\odot\hat{l}^i (i=1,2)$.
		\STATE Normalize the features: $M^i=\bar{M}^i/s^i$, where $s^i$ is the proportion of 1 in $\bar{M}^i (i=1,2)$.
		\STATE Return $M^1$, $M^2$.
	\end{algorithmic}
\end{algorithm}


\subsection{SDropDML}
\label{sec:SDropDML}
\begin{figure}
	\centering
	\includegraphics[width=0.95\linewidth]{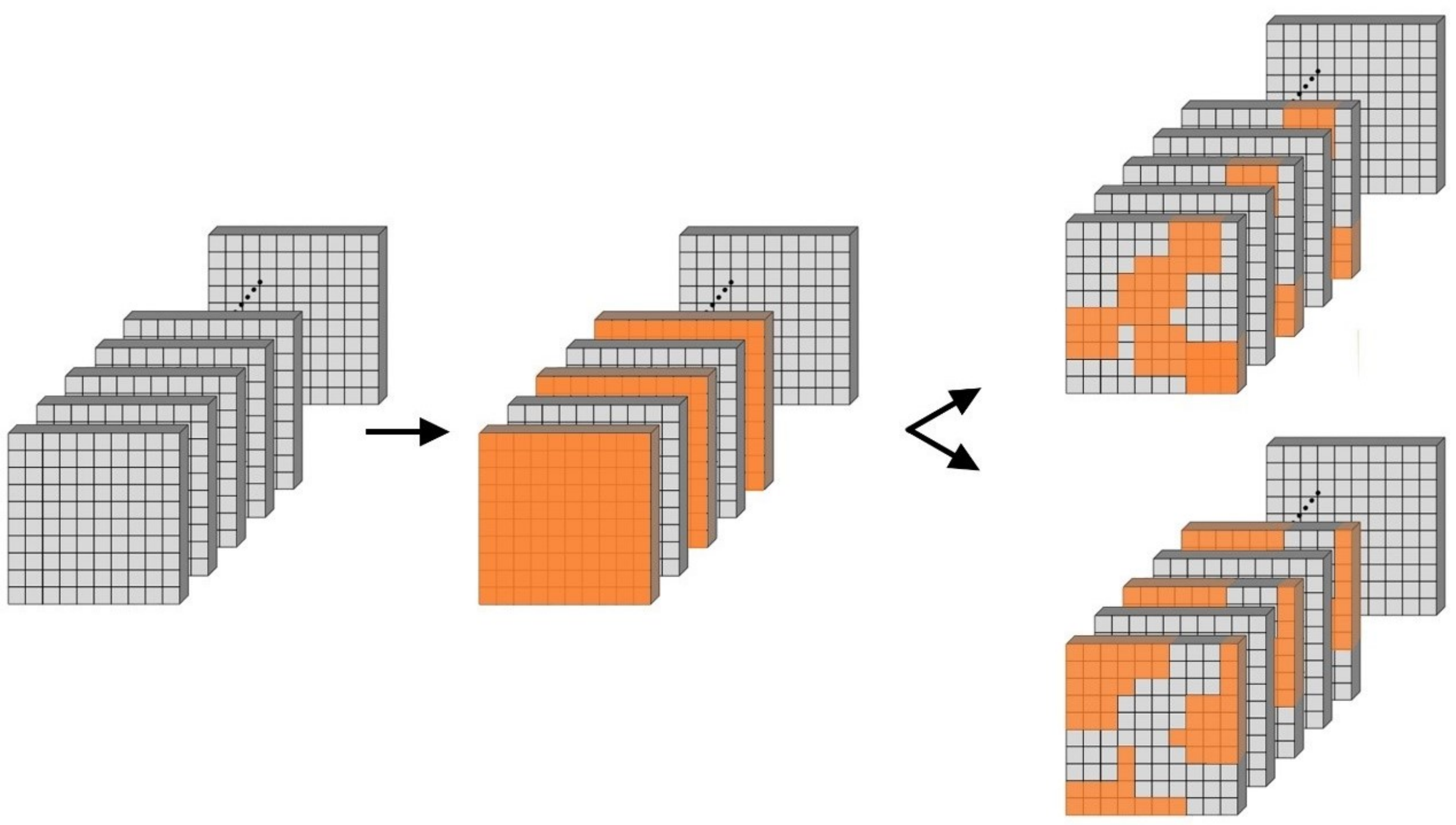}
	\caption{The overall framework of SDropDML.}
	\label{fig:SDropDML}
\end{figure}

We propose a strategy for constructing two sub models with complementary drop regions on the same channel, coined Spatial Dropout Deep Mutual Learning (SDropDML). Specifically, two sub models randomly drops the same channels of a feature map and then perform complementary DropBlock mask division with a probability of $0.5$ on each dropped channel. The overall framework of SDropDML is shown in Figure~\ref{fig:SDropDML}, where the yellow Blocks represent drop regions.

When calculating the complementary DropBlock mask, we need to first estimate the dropout probability $\gamma$ of block center based on the dropout probability $p=0.5$. Given one input $x\in\mathcal{R}^{m\times n\times c}$ and the size $b_{size}=2k+1$, when $m,n>2b_{size}$, the relationship between $p$ and $\gamma$ is as follows(detailed proof can be found in Sec.~\ref{sec:relationship}):
\begin{align}
	\label{eqn:relationship}
	p 
	&= (1-\frac{2k}{m})(1-\frac{2k}{n})\left( 1-(1-\gamma)^{(2k+1)^2}\right) \\ \nonumber
	&+4\left[ \frac{k^2}{mn}-\frac{1}{mn}\sum_{1\leq i,j\leq k}(1-\gamma)^{(k+i)(k+j)}\right] \\ \nonumber
	&+2\left( \frac{1}{m}+\frac{1}{n}-\frac{4k}{mn}\right) \left[ k-(1-\gamma)^{(2k+1)(k+1)}\frac{1-(1-\gamma)^{(2k+1)k}}{1-(1-\gamma)^{(2k+1)}}\right].\nonumber
\end{align}	

The details of SDropDML are given in Algorithm~\ref{al:SDropDML}.

\begin{algorithm}[htbp]
	\setstretch{1.28}
	\caption{SDropDML}
	\label{al:SDropDML}
	\textbf{Input}: $x\in\mathcal{R}^{m\times n\times c}$,  $b_{size}$, $p$. \\
	\textbf{Output}: Maskes sampling $M^1$, $M^2$.
	\begin{algorithmic}[1]	
		\STATE Randomly sample mask $l\in \mathcal{R}^c$: $l_k\sim\textit{Bernoulli}(1-p)$, $\forall k\in \{1,2,\ldots,c\}$.
		\STATE Estimate $\gamma$ by Eq.~(\ref{eqn:relationship}) with $p=0.5$, randomly sample mask $r\in \mathcal{R}^{m\times n}$: \\
			$r_{i,j}\sim\textit{Bernoulli}(1-\gamma)$, $\forall i\in \{1,2,\ldots,m\}$, $j\in\{1,2,\ldots,n\}$.
		\STATE  For each $1$ position $r_{i,j}$, create a spatial square mask $M^1\in\mathcal{R}^{m\times n}$ with the center being $r_{i,j}$, the width, height being $b_{size}$ and set all the values of $r$ in the square to be $1$.\\
			Calculate $M^2\in\mathcal{R}^{m\times n}$: $M^2=\textbf{1}_{M^1}-M^1$.
		\STATE Expand $M^1$, $M^2$ to $\hat{M}^1$, $\hat{M}^2\in\mathcal{R}^{m\times n\times c}$: $\hat{M}^1_{i,j,k}=M^1_{i,j}$, $\hat{M}^2_{i,j,k}=M^2_{i,j}$.\\
			Expand $l$ to $\hat{l}\in\mathcal{R}^{m\times n\times c}$: $\hat{l}_{i,j,k}=l_k$, \\
			$\forall i\in \{1,2,\ldots,m\}$, $j\in \{1,2,\ldots,n\}$, $k\in \{1,2,\ldots,c\}$.
		\STATE Calculate: $\bar{M}^i=\textbf{1}_x-\hat{l}\odot\hat{M}^i (i=1,2)$.
		\STATE Normalize the features: $M^i=\bar{M}^i/s^i$, where $s^i$ is the proportion of $1$ in $\bar{M}^i (i=1,2)$.
		\STATE Return $M^1$, $M^2$.
	\end{algorithmic}
\end{algorithm}


\subsection{A the operational relationship between $p$ and $\gamma$}
\label{sec:relationship}
In this section, we provide the derivation of Eq.~\ref{eqn:gamma1} -~\ref{eqn:relationship} about the operational relationship between the dropout probability $p$ of each activation unit and the dropout probability $\gamma$ of block center.

\begin{theorem}%
	Given one input $x\in\mathcal{R}^{m\times n\times c}$ and the size $b_{size}=2k+1$ of the block to be dropped, when $m,n>2b_{size}$, the operational relationship between the dropout probability $p$ of each activation unit and the dropout probability $\gamma$ of block center can be described as:
	\begin{align}
		p \nonumber
		&= (1-\frac{2k}{m})(1-\frac{2k}{n})\left( 1-(1-\gamma)^{(2k+1)^2}\right) \\ \nonumber
		&+4\left[ \frac{k^2}{mn}-\frac{1}{mn}\sum_{1\leq i,j\leq k}(1-\gamma)^{(k+i)(k+j)}\right] \\ \nonumber
		&+2\left( \frac{1}{m}+\frac{1}{n}-\frac{4k}{mn}\right) \left[ k-(1-\gamma)^{(2k+1)(k+1)}\frac{1-(1-\gamma)^{(2k+1)k}}{1-(1-\gamma)^{(2k+1)}}\right]\nonumber
	\end{align}	
\end{theorem}

Proof: Since the expansion principle of mask sampling in each channel is the same, we only consider the relationship between $p$ and $\gamma$ on $x\in\mathcal{R}^{m\times n}$.

Randomly sample mask  $r\in\mathcal{R}^{m\times n}$: $r_{i,j}\sim\textit{Bernoulli}(1-\gamma)$. For each $1$ position $r_{i,j}$, create a spatial square mask $M\in\mathcal{R}^{m\times n}$ with the center being $r_{i,j}$, the width, height being $b_{size}$ and set all the values of $r$ in the square to be $1$. 

\begin{figure}
	\centering
	\includegraphics[width=0.39\linewidth]{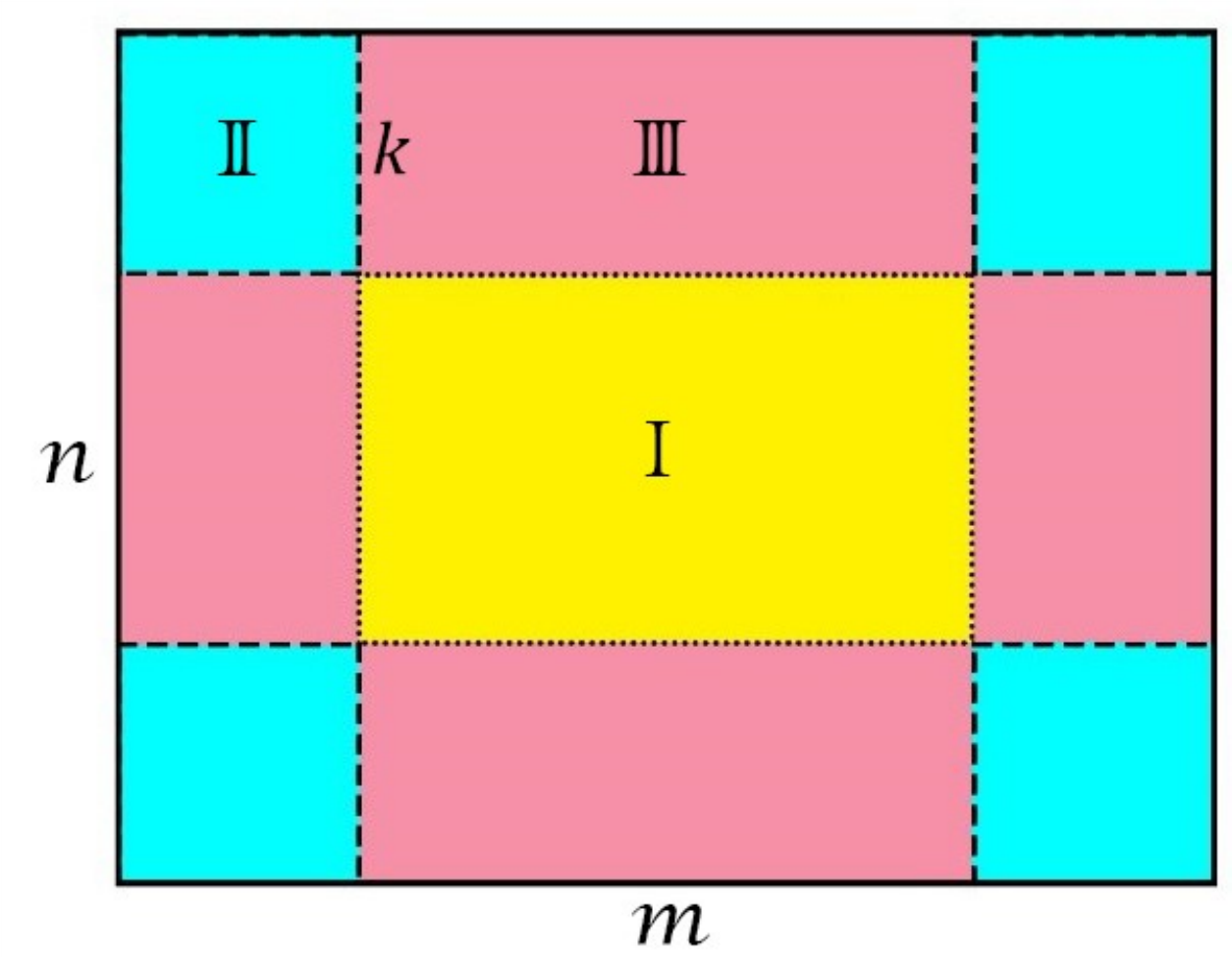}
	\caption{Partition of the mask $M\in\mathcal{R}^{m\times n}$.}
	\label{fig:partition}
\end{figure}

Here, the dropout probability for each activation unit in $M$ is $p$:
\begin{equation}
	p([M]_{i,j}=1\mid1\leq i\leq m,1\leq j\leq n)=p,
\end{equation}
We can calculate the expectation of the number of dropped activation units in $M$:
\begin{equation}
	\label{eqn:pmn}
	\mathbb{E}\left( \sum_{i,j}[M]_{i,j}\right) =\sum_{1\leq i\leq m,1\leq j\leq n}p([M]_{i,j}=1)=pmn,
\end{equation}
In order to express the dropout probability of $M$ with $\gamma$, we divide the mask $M$ into $3$ parts as shown in Figure~\ref{fig:partition}. 

In part \uppercase\expandafter{\romannumeral1}, the probability that the value of each activation unit equal to $1$ can be expressed as:
\begin{align}
	\label{eqn:p1}
	& p\left([M]_{i,j}=1\mid k+1\leq i\leq m-k,k+1\leq j\leq n-k\right)  \\  \nonumber
	= & p\left( \max_{-k+1\leq t,q\leq k+1}r_{i-1+t,j-1+q}=1\mid k+1\leq i\leq m-k,k+1\leq j\leq n-k\right)   \\ \nonumber
	= & 1-p\left( r_{i-1+t,j-1+q}=0\mid \forall -k+1\leq t,q\leq k+1\right)  \\ \nonumber
	= & 1-(1-\gamma)^{(2k+1)^2}. \nonumber
\end{align}
According to Eq.\ref{eqn:p1}, we have:
\begin{align}
	\label{eqn:e1}
	& \mathbb{E}\left(\sum_{i,j}[M]_{i,j}\mid [M]_{i,j} \in  \rm{\uppercase\expandafter{\romannumeral1}} \right)  \\  \nonumber
	= & \sum_{k+1\leq i\leq m-k,k+1\leq j\leq n-k}p\left([M]_{i,j}=1\right)   \\ \nonumber
	= & \sum_{k+1\leq i\leq m-k,k+1\leq j\leq n-k}\left(1-(1-\gamma)^{(2k+1)^2} \right)   \\ \nonumber
	= & (m-2k)(n-2k)\left( 1-(1-\gamma)^{(2k+1)^2}\right). \nonumber
\end{align}

In part \uppercase\expandafter{\romannumeral2}, due to symmetry, only the region $1\leq i,j\leq k$ needs to be considered. Then we have:
\begin{align}
	\label{eqn:p2}
	& p\left([M]_{i,j}=1\mid 1\leq i,j\leq k\right)  \\  \nonumber
	= & p\left( \max_{-k+1\leq t,q\leq k+1}r_{i-1+t,j-1+q}=1\mid 1\leq i,j\leq k\right)   \\ \nonumber
	= & 1-p\left( r_{i-1+t,j-1+q}=0\mid \forall 2-i\leq t\leq k+1, 2-j\leq q\leq k+1\right)  \\ \nonumber
	= & 1-(1-\gamma)^{(k+i)(k+j)}. \nonumber
\end{align}
The expectation of the number of dropped activation units in part \uppercase\expandafter{\romannumeral2} can be calculated as:
\begin{align}
	\label{eqn:e2}
	& \mathbb{E}\left(\sum_{i,j}[M]_{i,j}\mid [M]_{i,j} \in  \rm{\uppercase\expandafter{\romannumeral2}} \right)  \\  \nonumber
	= & 4\sum_{1\leq i,j\leq k}p\left([M]_{i,j}=1\right)   \\ \nonumber
	= & 4\sum_{1\leq i,j\leq k}\left(1-(1-\gamma)^{(k+i)(k+j)} \right)   \\ \nonumber
	= & 4\left[k^2- \sum_{1\leq i,j\leq k}(1-\gamma)^{(k+i)(k+j)} \right]. \nonumber
\end{align}
In part \uppercase\expandafter{\romannumeral3}, due to symmetry, only the regions $k+1\leq i\leq m-k$, $1\leq j\leq k$ and $1\leq i\leq k$, $k+1\leq j\leq n-k$ needs to be considered. The probability that the value of each activation unit in the regions $k+1\leq i\leq m-k$, $1\leq j\leq k$ equal to $1$ can be expressed as:
\begin{align}
	\label{eqn:p31}
	& p\left([M]_{i,j}=1\mid k+1\leq i\leq m-k,1\leq j\leq k\right)  \\  \nonumber
	= & p\left( \max_{-k+1\leq t,q\leq k+1}r_{i-1+t,j-1+q}=1\mid k+1\leq i\leq m-k,1\leq j\leq k\right)   \\ \nonumber
	= & 1-p\left( r_{i-1+t,j-1+q}=0\mid \forall -k+1\leq t\leq k+1, 2-j\leq q\leq k+1\right)  \\ \nonumber
	= & 1-(1-\gamma)^{(2k+1)(k+j)}. \nonumber
\end{align}
According to Eq~\ref{eqn:p31}, we have:
\begin{align}
	\nonumber
	& \mathbb{E}\left(\sum_{i,j}[M]_{i,j}\mid k+1\leq i\leq m-k,1\leq j\leq k \right)  \\  \nonumber
	= & \sum_{k+1\leq i\leq m-k,1\leq j\leq k}p\left([M]_{i,j}=1\right)   \\ \nonumber
	= &\sum_{k+1\leq i\leq m-k,1\leq j\leq k}\left(1-(1-\gamma)^{(2k+1)(k+j)} \right)   \\ \nonumber
	= & (m-2k)\sum_{1\leq j\leq k}\left(1-(1-\gamma)^{(2k+1)(k+j)} \right)\\ \nonumber 
	= & (m-2k)\left[ k-(1-\gamma)^{(2k+1)(k+1)}\frac{1-(1-\gamma)^{(2k+1)k}}{1-(1-\gamma)^{(2k+1)}}\right]. \nonumber
\end{align}
In the same way, the probability that the value of each activation unit in the regions $1\leq i\leq k$, $k+1\leq j\leq n-k$ equal to $1$ can be expressed as:
\begin{align}
	\label{eqn:p32}
	& p\left([M]_{i,j}=1\mid 1\leq i\leq k,k+1\leq j\leq n-k\right)  \\  \nonumber
	= & p\left( \max_{-k+1\leq t,q\leq k+1}r_{i-1+t,j-1+q}=1\mid 1\leq i\leq k,k+1\leq j\leq n-k\right)   \\ \nonumber
	= & 1-p\left( r_{i-1+t,j-1+q}=0\mid \forall 2-i\leq t\leq k+1,-k+1\leq q\leq k+1\right)  \\ \nonumber
	= & 1-(1-\gamma)^{(k+i)(2k+1)}. \nonumber
\end{align}
The corresponding expectation can be calculated as:
\begin{align}
	\nonumber
	& \mathbb{E}\left(\sum_{i,j}[M]_{i,j}\mid 1\leq i\leq k,k+1\leq j\leq n-k \right)  \\  \nonumber
	= & \sum_{1\leq i\leq k,k+1\leq j\leq n-k}p\left([M]_{i,j}=1\right)   \\ \nonumber
	= &\sum_{1\leq i\leq k,k+1\leq j\leq n-k}\left(1-(1-\gamma)^{(k+i)(2k+1)} \right)   \\ \nonumber
	= & (n-2k)\left[ k-(1-\gamma)^{(2k+1)(k+1)}\frac{1-(1-\gamma)^{(2k+1)k}}{1-(1-\gamma)^{(2k+1)}}\right]. \nonumber
\end{align}
The expectation of the number of dropped activation units in part \uppercase\expandafter{\romannumeral3} can be calculated as:
\begin{align}
	\label{eqn:e3}
	& \mathbb{E}\left(\sum_{i,j}[M]_{i,j}\mid [M]_{i,j} \in  \rm{\uppercase\expandafter{\romannumeral3}} \right)  \\  \nonumber
	= & 2\mathbb{E}\left(\sum_{i,j}[M]_{i,j}\mid k+1\leq i\leq m-k,1\leq j\leq k \right)  \\  \nonumber
	+ &2\mathbb{E}\left(\sum_{i,j}[M]_{i,j}\mid 1\leq i\leq k,k+1\leq j\leq n-k \right)  \\  \nonumber
	= & 2(m+n-4k)\left[ k-(1-\gamma)^{(2k+1)(k+1)}\frac{1-(1-\gamma)^{(2k+1)k}}{1-(1-\gamma)^{(2k+1)}}\right]. \nonumber
\end{align}
Combined with Eq.~\ref{eqn:pmn}, Eq.~\ref{eqn:e1}, Eq.~\ref{eqn:e2} and Eq.~\ref{eqn:e3}, we have:
	\begin{align}
	pmn \nonumber
	&= (m-2k)(n-2k)\left( 1-(1-\gamma)^{(2k+1)^2}\right) \\ \nonumber
	&+4\left[k^2-\sum_{1\leq i,j\leq k}(1-\gamma)^{(k+i)(k+j)}\right] \\ \nonumber
	&+2\left(m+n-4k\right) \left[ k-(1-\gamma)^{(2k+1)(k+1)}\frac{1-(1-\gamma)^{(2k+1)k}}{1-(1-\gamma)^{(2k+1)}}\right].\nonumber
\end{align}	
Divided by $mn$, we can get the result. $\hfill\square$

Particularly, when we use a sample mask $r$ to generate a spatial square mask $M$ without considering marginal loss, we have:
\begin{equation}
	\label{eqn:pmax}
	p=1-(1-\gamma)^{b_{size}^2}.
\end{equation}
If we only sample mask $r$ in part \uppercase\expandafter{\romannumeral3} in Figure~\ref{fig:partition}, we have:
\begin{equation} 
	\nonumber
	pmn=(m-b_{size}+1)(n-b_{size}+1)\left( 1-(1-\gamma)^{b_{size}^2}\right),
\end{equation}
that is: 
\begin{equation}
	\label{eqn:pmin}
	p=\frac{(m-b_{size}+1)(n-b_{size}+1)\left( 1-(1-\gamma)^{b_{size}^2}\right)}{mn}.
\end{equation}
For the same $\gamma$, we denote the probabilities calculated by Eq.~\ref{eqn:pmax} and Eq.~\ref{eqn:pmin} as $p_1$ and $p_1$ respectively, and then $p$ obtained by Eq.~\ref{eqn:relationship} satisfies:
\begin{equation} 
	\label{eqn:order}
	p_2\leq p\leq p_1.
\end{equation}
Suppose the value of $\gamma$ is small, according to the McLaughlin expansion, we have:
\begin{equation} 
	1-(1-\gamma)^{b_{size}^2}\sim b_{size}^2\gamma.
\end{equation}
Combined with Eq.~\ref{eqn:pmax} and Eq.~\ref{eqn:pmin}, we can easily get Eq.~\ref{eqn:gamma1} and Eq.~\ref{eqn:gamma2}.

In addition, we need to estimate $\gamma$ by Eq.~\ref{eqn:relationship} with $p=0.5$ in Algorithm~\ref{al:SDropDML}. In fact, $\gamma$ increases monotonously with the increase of $p$ from $0$ to $1$. Here, we calculate the initial values of $\gamma$ through Eq.~\ref{eqn:gamma1} and Eq.~\ref{eqn:gamma2} with $p=0.5$ and then adjustment $\gamma$ according to Eq.~\ref{eqn:relationship} and Monotonicity Eq.~\ref{eqn:order}.

\section{Experiments}
\label{sec:experiments}
	
\subsection{Experimental setup}
\noindent{\bf Datasets.} We demonstrate the effectiveness of R-Block on various  image classification datasets: CIFAR-10, CIFAR-100~\cite{Krizhevsky2009LearningML} and TinyImageNet~\cite{Deng2009ImageNetAL}. The CIFAR-10 and CIFAR-100 datasets consist of $32 \times 32$ color images containing objects from $10$ and $100$ classes respectively. Both are split into a 50,000-image train set and a 10,000-image test set. The TinyImageNet dataset contains 120,000 $64 \times 64$ colour images of $200$ object classes.

\noindent{\bf Network architecture.} We demonstrate our method on ResNet-18, ResNet-34~\cite{He2015DeepRL} and VGG16~\cite{Simonyan2014VeryDC}. We modify the first convolutional layer of ResNet-18 with kernel size $3 \times 3$, strides $1$ and padding $1$, instead of the kernel size $7 \times 7$, strides $2$ and padding $3$~\cite{He2016IdentityMI}.

\noindent{\bf Implementation Details.} We use stochastic gradient descents (SGD) with a momentum of $0.9$, an initial rate of $0.1$, weight decay of $5e-4$ and batch size of $128$. The learning rate is decayed by the factor of $0.1$ at $75$, $130$ and $180$ epochs for all datasets. The total epoch is set as $200$, the temperature $T$ is $3$ and the coefficient weight $\alpha$ is $0.1$. We set the dropout probability parameter $p$ as $0.2$ and the block size $b_{size}$ as $3$. Since baselines are usually overfitted for the longer training scheme and have lower validation accuracy at the end of training, we report the highest validation accuracy over the full training course for fair comparison.

\subsection{Classification accuracy}
We measure the top-1 ($\%$) accuracy of R-Block by comparing with Dropout, SpatialDropout and DropBlock on various datasets and model architectures. The results are presented in Table~\ref{tab:accuracy1} and Table~\ref{tab:accuracy2}. These results imply that R-Block induces better classification performance than others. For example, R-Block(BDropDML) improves the top-1 ($\%$) accuracy of Baseline from $71.28\%$ to $72.35\%$ under the CIFAR-100 dataset and ResNet-18. We also observe that the top-1 ($\%$) accuracy of Dropout with greater dropout probability tends to be worse than Baseline, which perhaps caused by insufficient training due to too many dropped activation unit.
 \begin{table}[!t]
		\centering
		\renewcommand\arraystretch{1.1} 
		\scalebox{0.92}{
			\begin{tabular}{l l l c c c}
				\toprule
			\textbf{Model} & \textbf{Method} & \textbf{\textit{p}} & \textbf{CIFAR-10} & \textbf{CIFAR-100}  \\
				\midrule
				\multirow{6}{*}{ResNet-18} 
				& Baseline & 0 & 92.98 & 71.28 \\
				& Dropout & 0.5 & 92.71 & 70.47 \\
				& SpatialDropout & 0.1 & 93.02 & 71.30 \\
				& DropBlock & 0.1 & 93.02 & 71.38 \\
				& \textbf{R-Block(BDropDML)} & 0.2 & \textbf{\textcolor{blue}{93.49}} & \textbf{72.35} \\
				& \textbf{R-Block(SDropDML)} & 0.2 & \textbf{93.56} & \textbf{\textcolor{blue}{72.08}} \\
				\midrule
				\multirow{6}{*}{VGG16} 
				& Baseline & 0 & 93.64 & 73.20 \\
				& Dropout & 0.5 & 93.56 & 72.93 \\
				& SpatialDropout & 0.1 & 93.75 & 73.08 \\
				& DropBlock & 0.1 & 93.71 & 73.27 \\
				& \textbf{R-Block(BDropDML)} & 0.2 & \textbf{94.24} & \textbf{\textcolor{blue}{73.80}}\\
				& \textbf{R-Block(SDropDML)} & 0.2 & \textbf{\textcolor{blue}{94.21}} & \textbf{73.98} \\
				\bottomrule
		\end{tabular}}
		\\[0.2cm]
		\caption{
			Top-1 ($\%$) accuracy on the CIFAR datasets and different model architectures. The best and second-best results are indicated in black bold and blue bold respectively.
		}
		\label{tab:accuracy1}
 \end{table}

\begin{table}[!t]
		\centering
		\renewcommand\arraystretch{1.1} 
		\scalebox{0.92}{
		\begin{tabular}{l l c}
			\toprule
			\textbf{Method} & \textbf{\textit{p}} & \textbf{TinyImageNet}  \\
			\midrule
			 Baseline & 0 & 42.96 \\
			 SpatialDropout & 0-0.1 & 43.02 \\
			 DropBlock & 0-0.1 & 43.07 \\
			 \textbf{R-Block(BDropDML)} & 0-0.2 & \textbf{43.37} \\
			 \textbf{R-Block(SDropDML)} & 0-0.2 & \textbf{\textcolor{blue}{43.35}} \\
			\bottomrule
		\end{tabular}}
		\\[0.2cm]
		\caption{
			Top-1 ($\%$) accuracy on theTinyImageNet dataset and ResNet-34. The best and second-best results are indicated in black bold and blue bold respectively. In this experiments, we use a linear scheme of increasing the value of dropout probability. This linear scheme is similar to ScheduledDropPath~\cite{Zoph2017LearningTA}.
		}
		\label{tab:accuracy2}
\end{table}

\subsection{Methods with different sub models}

\begin{figure}
	\centering
	\includegraphics[width=0.98\linewidth]{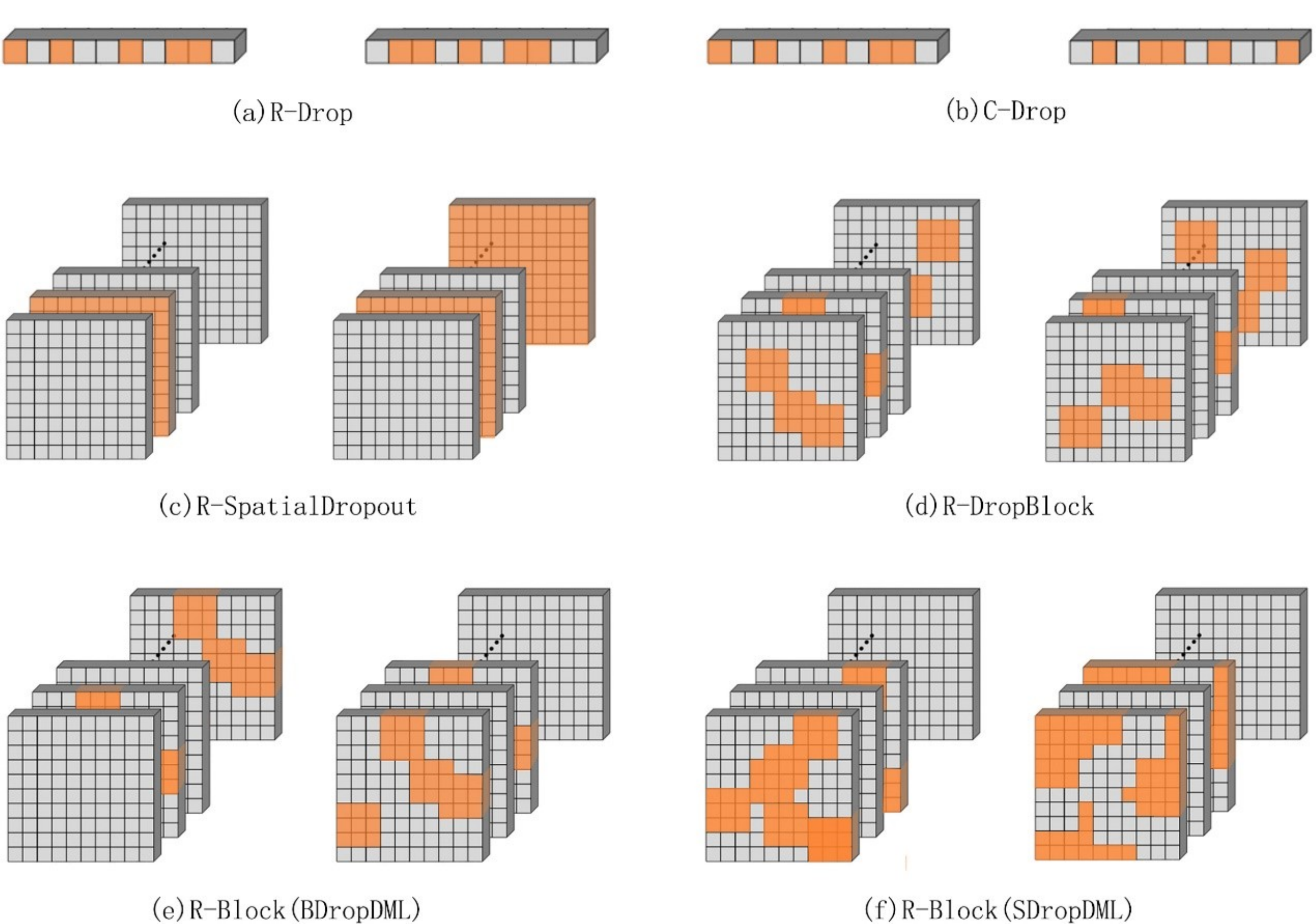}
	\caption{Different approaches to construct sub models.}
	\label{fig:submodels}
\end{figure}

We provide six approaches to construct sub models as shown in Figure~\ref{fig:submodels} (a)-(f). The details of these approaches are listed below:

\begin{enumerate}[(a)]
	\item This approach is to use Dropout in fully connected layers twice randomly, corresponding to R-Drop~\cite{Liang2021RDropRD}.
	\item This approach is inspired by R-Drop and two sub models have complementary drop regions in fully connected layers. We denote it as C-Drop.
	\item This approach is to use SpatialDropout twice randomly and we denote it as R-SpatialDropout.
	\item This approach is to use DropBlock twice randomly and we denote it as R-DropBlock.
	\item This approach is BDropDML in Sec.~\ref{sec:BDropDML}, corresponding to ours method R-Block(BDropDML).
	\item This approach is SDropDML in Sec.~\ref{sec:SDropDML}, corresponding to ours method R-Block(SDropDML).
\end{enumerate}

 \begin{table}[!t]
		\centering
		\renewcommand\arraystretch{1.1} 
		\scalebox{0.92}{
		\begin{tabular}{l l | c c c c c}
			\toprule
			\multirow{2}{*}{\textbf{Method}} & \multirow{2}{*}{\textbf{\textit{p}}} & \multicolumn{5}{c}{\textbf{Training stages}} \\
			&  & \textbf{20$\%$} & \textbf{40$\%$} & \textbf{60$\%$} & \textbf{80$\%$} & \textbf{100$\%$} \\
			\midrule
			 R-Drop & 0.5 & 54.93 & 68.88 & 68.99 & 70.81 & 71.17 \\
			 C-Drop & 0.5 & 54.60 & 69.50 & 69.70 & 71.03 & 71.30 \\
			 R-SpatialDropout & 0.1 & 55.80 & 69.83 & 69.98 & 71.41 & 71.86 \\
			 R-DropBlock & 0.1 & 56.20 & 69.81 & 70.30 & 71.15 & 71.60 \\
			 \textbf{R-Block(BDropDML)} & 0.2 & \textbf{\textcolor{blue}{56.83}} & \textbf{70.21} & \textbf{70.83} & \textbf{71.98}& \textbf{72.35}\\
			 \textbf{R-Block(SDropDML)} & 0.2 & \textbf{58.19} & \textbf{\textcolor{blue}{70.15}} & \textbf{\textcolor{blue}{70.31}} & \textbf{\textcolor{blue}{71.49}} & \textbf{\textcolor{blue}{72.08}} \\
			\bottomrule
		\end{tabular}}
		\\[0.2cm]
		\caption{
			Top-1 ($\%$) accuracy of various methods with different sub models at different training stages on CIFAR-100 and ResNet-18. Each stage adopts the optimal model of the current stage. The best and second-best results are indicated in black bold and blue bold respectively.
		}
		\label{tab:accuracy3}
\end{table}

We report the performance of R-Drop, C-Drop, R-SpatialDropout, R-DropBlock, R-Block(BDropDML) and R-Block(SDropDML) corresponding to (a)-(f) respectively at different training stages in Table~\ref{tab:accuracy3}. The results show that our R-Block performs the best. These also show that methods using sub models with different drop regions outperform methods using sub models with random drop regions. Sub models with complementary drop regions enable each activation unit to update during training. At the same time, since the drop regions of two sub models in BDropDML and SDropDML are completely different, the difference in semantic information of the feature map is extracted to the greatest extent, which greatly improves the regularization efficiency of the sub models.

\section{Conclusion and Future Work}

In this paper, we propose R-Block to regularize training CNNs. R-Block is a form of structured dropout and can reduce the inconsistency of the model structure between training and inference phases by minimizing the mutual learning losses on the outputs of two sub models for each data sample in training. We introduce BDropDML and SDropDML to construct sub models with different drop regions. Our experiments on CIFAR and TinyImageNet demonstrate that R-Block performs better than other existing structured dropout variants and methods using sub models with different drop regions outperform methods using sub models with random drop regions. Our next step is to consider adding an attention mechanism on feature maps and dynamically segment the drop regions to construct sub models more reasonably.

\bibliographystyle{mybst}
\bibliography{neurips_2021}

\begin{thebibliography}{81}
\providecommand{\natexlab}[1]{#1}
\providecommand{\url}[1]{\texttt{#1}}
\expandafter\ifx\csname urlstyle\endcsname\relax
  \providecommand{\doi}[1]{doi: #1}\else
  \providecommand{\doi}{doi: \begingroup \urlstyle{rm}\Url}\fi

\bibitem[Aghajanyan et~al.(2020)Aghajanyan, Shrivastava, Gupta, Goyal,
  Zettlemoyer, and Gupta]{aghajanyan2020better}
Armen Aghajanyan, Akshat Shrivastava, Anchit Gupta, Naman Goyal, Luke
  Zettlemoyer, and Sonal Gupta.
\newblock Better fine-tuning by reducing representational collapse.
\newblock \emph{arXiv preprint arXiv:2008.03156}, 2020.

\bibitem[Allen-Zhu \& Li(2020)Allen-Zhu and Li]{allen2020towards}
Zeyuan Allen-Zhu and Yuanzhi Li.
\newblock Towards understanding ensemble, knowledge distillation and
  self-distillation in deep learning.
\newblock \emph{arXiv preprint arXiv:2012.09816}, 2020.

\bibitem[Ba et~al.(2016)Ba, Kiros, and Hinton]{ba2016layer}
Jimmy~Lei Ba, Jamie~Ryan Kiros, and Geoffrey~E Hinton.
\newblock Layer normalization.
\newblock \emph{arXiv preprint arXiv:1607.06450}, 2016.

\bibitem[Ba \& Frey(2013)Ba and Frey]{ba2013adaptive}
Lei~Jimmy Ba and Brendan Frey.
\newblock Adaptive dropout for training deep neural networks.
\newblock In \emph{Proceedings of the 26th International Conference on Neural
  Information Processing Systems-Volume 2}, pp.\  3084--3092, 2013.

\bibitem[Baevski \& Auli(2018)Baevski and Auli]{baevski2018adaptive}
Alexei Baevski and Michael Auli.
\newblock Adaptive input representations for neural language modeling.
\newblock In \emph{International Conference on Learning Representations}, 2018.

\bibitem[Brown et~al.(2020)Brown, Mann, Ryder, Subbiah, Kaplan, Dhariwal,
  Neelakantan, Shyam, Sastry, Askell, et~al.]{brown2020language}
Tom~B Brown, Benjamin Mann, Nick Ryder, Melanie Subbiah, Jared Kaplan, Prafulla
  Dhariwal, Arvind Neelakantan, Pranav Shyam, Girish Sastry, Amanda Askell,
  et~al.
\newblock Language models are few-shot learners.
\newblock \emph{arXiv preprint arXiv:2005.14165}, 2020.

\bibitem[Clark et~al.(2019)Clark, Luong, Le, and Manning]{clark2019electra}
Kevin Clark, Minh-Thang Luong, Quoc~V Le, and Christopher~D Manning.
\newblock Electra: Pre-training text encoders as discriminators rather than
  generators.
\newblock In \emph{International Conference on Learning Representations}, 2019.

\bibitem[Deng et~al.(2009)Deng, Dong, Socher, Li, Li, and
  Fei-Fei]{deng2009imagenet}
Jia Deng, Wei Dong, Richard Socher, Li-Jia Li, Kai Li, and Li~Fei-Fei.
\newblock Imagenet: A large-scale hierarchical image database.
\newblock In \emph{2009 IEEE conference on computer vision and pattern
  recognition}, pp.\  248--255. Ieee, 2009.

\bibitem[Devlin et~al.(2019)Devlin, Chang, Lee, and Toutanova]{devlin2019bert}
Jacob Devlin, Ming-Wei Chang, Kenton Lee, and Kristina Toutanova.
\newblock Bert: Pre-training of deep bidirectional transformers for language
  understanding.
\newblock In \emph{Proceedings of the 2019 Conference of the North American
  Chapter of the Association for Computational Linguistics: Human Language
  Technologies, Volume 1 (Long and Short Papers)}, pp.\  4171--4186, 2019.

\bibitem[DeVries \& Taylor(2017)DeVries and Taylor]{devries2017improved}
Terrance DeVries and Graham~W Taylor.
\newblock Improved regularization of convolutional neural networks with cutout.
\newblock \emph{arXiv preprint arXiv:1708.04552}, 2017.

\bibitem[Dosovitskiy et~al.(2021)Dosovitskiy, Beyer, Kolesnikov, Weissenborn,
  Zhai, Unterthiner, Dehghani, Minderer, Heigold, Gelly, Uszkoreit, and
  Houlsby]{dosovitskiy2021an}
Alexey Dosovitskiy, Lucas Beyer, Alexander Kolesnikov, Dirk Weissenborn,
  Xiaohua Zhai, Thomas Unterthiner, Mostafa Dehghani, Matthias Minderer, Georg
  Heigold, Sylvain Gelly, Jakob Uszkoreit, and Neil Houlsby.
\newblock An image is worth 16x16 words: Transformers for image recognition at
  scale.
\newblock In \emph{International Conference on Learning Representations}, 2021.

\bibitem[Erhan et~al.(2009)Erhan, Manzagol, Bengio, Bengio, and
  Vincent]{erhan2009difficulty}
Dumitru Erhan, Pierre-Antoine Manzagol, Yoshua Bengio, Samy Bengio, and Pascal
  Vincent.
\newblock The difficulty of training deep architectures and the effect of
  unsupervised pre-training.
\newblock In \emph{Artificial Intelligence and Statistics}, pp.\  153--160.
  PMLR, 2009.

\bibitem[Fang et~al.(2021)Fang, Wang, Wang, Zhang, Yang, and Liu]{fang2021seed}
Zhiyuan Fang, Jianfeng Wang, Lijuan Wang, Lei Zhang, Yezhou Yang, and Zicheng
  Liu.
\newblock Seed: Self-supervised distillation for visual representation.
\newblock \emph{International Conference on Learning Representations}, 2021.

\bibitem[Fedus et~al.(2021)Fedus, Zoph, and Shazeer]{fedus2021switch}
William Fedus, Barret Zoph, and Noam Shazeer.
\newblock Switch transformers: Scaling to trillion parameter models with simple
  and efficient sparsity.
\newblock \emph{arXiv preprint arXiv:2101.03961}, 2021.

\bibitem[Furlanello et~al.(2018)Furlanello, Lipton, Tschannen, Itti, and
  Anandkumar]{furlanello2018born}
Tommaso Furlanello, Zachary Lipton, Michael Tschannen, Laurent Itti, and Anima
  Anandkumar.
\newblock Born again neural networks.
\newblock In \emph{International Conference on Machine Learning}, pp.\
  1607--1616. PMLR, 2018.

\bibitem[Gal \& Ghahramani(2016{\natexlab{a}})Gal and
  Ghahramani]{gal2016dropout}
Yarin Gal and Zoubin Ghahramani.
\newblock Dropout as a bayesian approximation: Representing model uncertainty
  in deep learning.
\newblock In \emph{international conference on machine learning}, pp.\
  1050--1059. PMLR, 2016{\natexlab{a}}.

\bibitem[Gal \& Ghahramani(2016{\natexlab{b}})Gal and
  Ghahramani]{gal2016theoretically}
Yarin Gal and Zoubin Ghahramani.
\newblock A theoretically grounded application of dropout in recurrent neural
  networks.
\newblock \emph{Advances in neural information processing systems},
  29:\penalty0 1019--1027, 2016{\natexlab{b}}.

\bibitem[Gao et~al.(2021)Gao, Yao, and Chen]{gao2021simcse}
Tianyu Gao, Xingcheng Yao, and Danqi Chen.
\newblock Simcse: Simple contrastive learning of sentence embeddings.
\newblock \emph{arXiv preprint arXiv:2104.08821}, 2021.

\bibitem[Gotmare et~al.(2019)Gotmare, Keskar, Xiong, and
  Socher]{gotmare2018closer}
Akhilesh Gotmare, Nitish~Shirish Keskar, Caiming Xiong, and Richard Socher.
\newblock A closer look at deep learning heuristics: Learning rate restarts,
  warmup and distillation.
\newblock In \emph{International Conference on Learning Representations}, 2019.

\bibitem[Hahn \& Choi(2019)Hahn and Choi]{hahn2019self}
Sangchul Hahn and Heeyoul Choi.
\newblock Self-knowledge distillation in natural language processing.
\newblock In \emph{Proceedings of the International Conference on Recent
  Advances in Natural Language Processing (RANLP 2019)}, pp.\  423--430, 2019.

\bibitem[He et~al.(2015)He, Zhang, Ren, and Sun]{he2015delving}
Kaiming He, Xiangyu Zhang, Shaoqing Ren, and Jian Sun.
\newblock Delving deep into rectifiers: Surpassing human-level performance on
  imagenet classification.
\newblock In \emph{Proceedings of the IEEE international conference on computer
  vision}, pp.\  1026--1034, 2015.

\bibitem[Hermann et~al.(2015)Hermann, Ko{\v{c}}isk{\`y}, Grefenstette,
  Espeholt, Kay, Suleyman, and Blunsom]{hermann2015teaching}
Karl~Moritz Hermann, Tom{\'a}{\v{s}} Ko{\v{c}}isk{\`y}, Edward Grefenstette,
  Lasse Espeholt, Will Kay, Mustafa Suleyman, and Phil Blunsom.
\newblock Teaching machines to read and comprehend.
\newblock \emph{arXiv preprint arXiv:1506.03340}, 2015.

\bibitem[Hinton et~al.(2015)Hinton, Vinyals, and Dean]{hinton2015distilling}
Geoffrey Hinton, Oriol Vinyals, and Jeff Dean.
\newblock Distilling the knowledge in a neural network.
\newblock \emph{arXiv preprint arXiv:1503.02531}, 2015.

\bibitem[Hinton et~al.(2012)Hinton, Srivastava, Krizhevsky, Sutskever, and
  Salakhutdinov]{hinton2012improving}
Geoffrey~E Hinton, Nitish Srivastava, Alex Krizhevsky, Ilya Sutskever, and
  Ruslan~R Salakhutdinov.
\newblock Improving neural networks by preventing co-adaptation of feature
  detectors.
\newblock \emph{arXiv preprint arXiv:1207.0580}, 2012.

\bibitem[Hochreiter \& Schmidhuber(1995)Hochreiter and
  Schmidhuber]{hochreiter1995simplifying}
Sepp Hochreiter and J{\"u}rgen Schmidhuber.
\newblock Simplifying neural nets by discovering flat minima.
\newblock In \emph{Advances in neural information processing systems}, pp.\
  529--536, 1995.

\bibitem[Huang et~al.(2018)Huang, Liu, Lang, Yu, Wang, and
  Li]{huang2018orthogonal}
Lei Huang, Xianglong Liu, Bo~Lang, Adams Yu, Yongliang Wang, and Bo~Li.
\newblock Orthogonal weight normalization: Solution to optimization over
  multiple dependent stiefel manifolds in deep neural networks.
\newblock In \emph{Proceedings of the AAAI Conference on Artificial
  Intelligence}, volume~32, 2018.

\bibitem[Ioffe \& Szegedy(2015)Ioffe and Szegedy]{ioffe2015batch}
Sergey Ioffe and Christian Szegedy.
\newblock Batch normalization: Accelerating deep network training by reducing
  internal covariate shift.
\newblock In \emph{International conference on machine learning}, pp.\
  448--456. PMLR, 2015.

\bibitem[Kang et~al.(2016)Kang, Li, and Tao]{kang2016shakeout}
Guoliang Kang, Jun Li, and Dacheng Tao.
\newblock Shakeout: A new regularized deep neural network training scheme.
\newblock In \emph{Proceedings of the AAAI Conference on Artificial
  Intelligence}, volume~30, 2016.

\bibitem[Kingma \& Ba(2014)Kingma and Ba]{kingma2014adam}
Diederik~P Kingma and Jimmy Ba.
\newblock Adam: A method for stochastic optimization.
\newblock \emph{arXiv preprint arXiv:1412.6980}, 2014.

\bibitem[Kolesnikov et~al.(2019)Kolesnikov, Beyer, Zhai, Puigcerver, Yung,
  Gelly, and Houlsby]{kolesnikov2019big}
Alexander Kolesnikov, Lucas Beyer, Xiaohua Zhai, Joan Puigcerver, Jessica Yung,
  Sylvain Gelly, and Neil Houlsby.
\newblock Big transfer (bit): General visual representation learning.
\newblock \emph{arXiv preprint arXiv:1912.11370}, 6\penalty0 (2):\penalty0 8,
  2019.

\bibitem[Krizhevsky et~al.(2012)Krizhevsky, Sutskever, and
  Hinton]{krizhevsky2012imagenet}
Alex Krizhevsky, Ilya Sutskever, and Geoffrey~E Hinton.
\newblock Imagenet classification with deep convolutional neural networks.
\newblock \emph{Advances in neural information processing systems},
  25:\penalty0 1097--1105, 2012.

\bibitem[Krizhevsky et~al.(2009)]{krizhevsky2009learning}
Alex Krizhevsky et~al.
\newblock Learning multiple layers of features from tiny images.
\newblock 2009.

\bibitem[Krogh \& Hertz(1992)Krogh and Hertz]{krogh1992simple}
Anders Krogh and John~A Hertz.
\newblock A simple weight decay can improve generalization.
\newblock In \emph{Advances in neural information processing systems}, pp.\
  950--957, 1992.

\bibitem[Labach et~al.(2019)Labach, Salehinejad, and Valaee]{labach2019survey}
Alex Labach, Hojjat Salehinejad, and Shahrokh Valaee.
\newblock Survey of dropout methods for deep neural networks.
\newblock \emph{arXiv preprint arXiv:1904.13310}, 2019.

\bibitem[Lewis et~al.(2020)Lewis, Liu, Goyal, Ghazvininejad, Mohamed, Levy,
  Stoyanov, and Zettlemoyer]{lewis2020bart}
Mike Lewis, Yinhan Liu, Naman Goyal, Marjan Ghazvininejad, Abdelrahman Mohamed,
  Omer Levy, Veselin Stoyanov, and Luke Zettlemoyer.
\newblock Bart: Denoising sequence-to-sequence pre-training for natural
  language generation, translation, and comprehension.
\newblock In \emph{Proceedings of the 58th Annual Meeting of the Association
  for Computational Linguistics}, pp.\  7871--7880, 2020.

\bibitem[Liang et~al.(2021)Liang, Hao, Shen, Zhou, Chen, Chen, and
  Carin]{liang2021mixkd}
Kevin~J Liang, Weituo Hao, Dinghan Shen, Yufan Zhou, Weizhu Chen, Changyou
  Chen, and Lawrence Carin.
\newblock Mixkd: Towards efficient distillation of large-scale language models.
\newblock \emph{International Conference on Learning Representations}, 2021.

\bibitem[Lin \& Hovy(2002)Lin and Hovy]{lin2002manual}
Chin-Yew Lin and Eduard Hovy.
\newblock Manual and automatic evaluation of summaries.
\newblock In \emph{Proceedings of the ACL-02 Workshop on Automatic
  Summarization}, pp.\  45--51, 2002.

\bibitem[Liu et~al.(2020)Liu, Duh, Liu, and Gao]{liu2020very}
Xiaodong Liu, Kevin Duh, Liyuan Liu, and Jianfeng Gao.
\newblock Very deep transformers for neural machine translation.
\newblock \emph{arXiv preprint arXiv:2008.07772}, 2020.

\bibitem[Liu et~al.(2019)Liu, Ott, Goyal, Du, Joshi, Chen, Levy, Lewis,
  Zettlemoyer, and Stoyanov]{liu2019roberta}
Yinhan Liu, Myle Ott, Naman Goyal, Jingfei Du, Mandar Joshi, Danqi Chen, Omer
  Levy, Mike Lewis, Luke Zettlemoyer, and Veselin Stoyanov.
\newblock Roberta: A robustly optimized bert pretraining approach.
\newblock \emph{arXiv preprint arXiv:1907.11692}, 2019.

\bibitem[Liu et~al.(2021)Liu, Lin, Cao, Hu, Wei, Zhang, Lin, and
  Guo]{liu2021swin}
Ze~Liu, Yutong Lin, Yue Cao, Han Hu, Yixuan Wei, Zheng Zhang, Stephen Lin, and
  Baining Guo.
\newblock Swin transformer: Hierarchical vision transformer using shifted
  windows.
\newblock \emph{arXiv preprint arXiv:2103.14030}, 2021.

\bibitem[Merity et~al.(2016)Merity, Xiong, Bradbury, and
  Socher]{merity2016pointer}
Stephen Merity, Caiming Xiong, James Bradbury, and Richard Socher.
\newblock Pointer sentinel mixture models.
\newblock \emph{arXiv preprint arXiv:1609.07843}, 2016.

\bibitem[Merity et~al.(2018)Merity, Keskar, and Socher]{merity2018regularizing}
Stephen Merity, Nitish~Shirish Keskar, and Richard Socher.
\newblock Regularizing and optimizing lstm language models.
\newblock In \emph{International Conference on Learning Representations}, 2018.

\bibitem[Mobahi et~al.(2020)Mobahi, Farajtabar, and Bartlett]{mobahi2020self}
Hossein Mobahi, Mehrdad Farajtabar, and Peter~L Bartlett.
\newblock Self-distillation amplifies regularization in hilbert space.
\newblock \emph{arXiv preprint arXiv:2002.05715}, 2020.

\bibitem[Molchanov et~al.(2017)Molchanov, Ashukha, and
  Vetrov]{molchanov2017variational}
Dmitry Molchanov, Arsenii Ashukha, and Dmitry Vetrov.
\newblock Variational dropout sparsifies deep neural networks.
\newblock In \emph{International Conference on Machine Learning}, pp.\
  2498--2507. PMLR, 2017.

\bibitem[Moradi et~al.(2020)Moradi, Berangi, and Minaei]{moradi2020survey}
Reza Moradi, Reza Berangi, and Behrouz Minaei.
\newblock A survey of regularization strategies for deep models.
\newblock \emph{Artificial Intelligence Review}, 53\penalty0 (6):\penalty0
  3947--3986, 2020.

\bibitem[Neklyudov et~al.(2017)Neklyudov, Molchanov, Ashukha, and
  Vetrov]{neklyudov2017structured}
Kirill Neklyudov, Dmitry Molchanov, Arsenii Ashukha, and Dmitry Vetrov.
\newblock Structured bayesian pruning via log-normal multiplicative noise.
\newblock In \emph{Proceedings of the 31st International Conference on Neural
  Information Processing Systems}, pp.\  6778--6787, 2017.

\bibitem[Nguyen et~al.(2020)Nguyen, Joty, Wu, and Aw]{nguyen2020data}
Xuan-Phi Nguyen, Shafiq Joty, Kui Wu, and Ai~Ti Aw.
\newblock Data diversification: A simple strategy for neural machine
  translation.
\newblock In \emph{Advances in Neural Information Processing Systems}, pp.\
  10018--10029, 2020.

\bibitem[Ott et~al.(2018)Ott, Edunov, Grangier, and Auli]{ott2018scaling}
Myle Ott, Sergey Edunov, David Grangier, and Michael Auli.
\newblock Scaling neural machine translation.
\newblock In \emph{Proceedings of the Third Conference on Machine Translation:
  Research Papers}, pp.\  1--9, 2018.

\bibitem[Ott et~al.(2019)Ott, Edunov, Baevski, Fan, Gross, Ng, Grangier, and
  Auli]{ott2019fairseq}
Myle Ott, Sergey Edunov, Alexei Baevski, Angela Fan, Sam Gross, Nathan Ng,
  David Grangier, and Michael Auli.
\newblock fairseq: A fast, extensible toolkit for sequence modeling.
\newblock In \emph{NAACL-HLT (Demonstrations)}, 2019.

\bibitem[Pham \& Le(2021)Pham and Le]{pham2021autodropout}
Hieu Pham and Quoc~V Le.
\newblock Autodropout: Learning dropout patterns to regularize deep networks.
\newblock \emph{arXiv preprint arXiv:2101.01761}, 2021.

\bibitem[Poole et~al.(2014)Poole, Sohl-Dickstein, and
  Ganguli]{poole2014analyzing}
Ben Poole, Jascha Sohl-Dickstein, and Surya Ganguli.
\newblock Analyzing noise in autoencoders and deep networks.
\newblock \emph{arXiv preprint arXiv:1406.1831}, 2014.

\bibitem[Post(2018)]{post2018call}
Matt Post.
\newblock A call for clarity in reporting bleu scores.
\newblock In \emph{Proceedings of the Third Conference on Machine Translation:
  Research Papers}. Association for Computational Linguistics, 2018.

\bibitem[Qi et~al.(2020)Qi, Yan, Gong, Liu, Duan, Chen, Zhang, and
  Zhou]{qi2020prophetnet}
Weizhen Qi, Yu~Yan, Yeyun Gong, Dayiheng Liu, Nan Duan, Jiusheng Chen, Ruofei
  Zhang, and Ming Zhou.
\newblock Prophetnet: Predicting future n-gram for sequence-to-sequence
  pre-training.
\newblock In \emph{Proceedings of the 2020 Conference on Empirical Methods in
  Natural Language Processing: Findings}, pp.\  2401--2410, 2020.

\bibitem[Radford et~al.()Radford, Narasimhan, Salimans, and
  Sutskever]{radford2018improving}
Alec Radford, Karthik Narasimhan, Tim Salimans, and Ilya Sutskever.
\newblock Improving language understanding by generative pre-training.

\bibitem[Radford et~al.(2019)Radford, Wu, Child, Luan, Amodei, and
  Sutskever]{radford2019language}
Alec Radford, Jeffrey Wu, Rewon Child, David Luan, Dario Amodei, and Ilya
  Sutskever.
\newblock Language models are unsupervised multitask learners.
\newblock \emph{OpenAI blog}, 1\penalty0 (8):\penalty0 9, 2019.

\bibitem[Salimans \& Kingma(2016)Salimans and Kingma]{salimans2016weight}
Tim Salimans and Diederik~P Kingma.
\newblock Weight normalization: a simple reparameterization to accelerate
  training of deep neural networks.
\newblock In \emph{Proceedings of the 30th International Conference on Neural
  Information Processing Systems}, pp.\  901--909, 2016.

\bibitem[Semeniuta et~al.(2016)Semeniuta, Severyn, and
  Barth]{semeniuta2016recurrent}
Stanislau Semeniuta, Aliaksei Severyn, and Erhardt Barth.
\newblock Recurrent dropout without memory loss.
\newblock In \emph{Proceedings of COLING 2016, the 26th International
  Conference on Computational Linguistics: Technical Papers}, pp.\  1757--1766,
  2016.

\bibitem[Sennrich et~al.(2016)Sennrich, Haddow, and Birch]{sennrich2016neural}
Rico Sennrich, Barry Haddow, and Alexandra Birch.
\newblock Neural machine translation of rare words with subword units.
\newblock In \emph{Proceedings of the 54th Annual Meeting of the Association
  for Computational Linguistics (Volume 1: Long Papers)}, pp.\  1715--1725,
  2016.

\bibitem[Shen et~al.(2020)Shen, Zheng, Shen, Qu, and Chen]{shen2020simple}
Dinghan Shen, Mingzhi Zheng, Yelong Shen, Yanru Qu, and Weizhu Chen.
\newblock A simple but tough-to-beat data augmentation approach for natural
  language understanding and generation.
\newblock \emph{arXiv preprint arXiv:2009.13818}, 2020.

\bibitem[Srivastava et~al.(2014)Srivastava, Hinton, Krizhevsky, Sutskever, and
  Salakhutdinov]{srivastava2014dropout}
Nitish Srivastava, Geoffrey Hinton, Alex Krizhevsky, Ilya Sutskever, and Ruslan
  Salakhutdinov.
\newblock Dropout: a simple way to prevent neural networks from overfitting.
\newblock \emph{The journal of machine learning research}, 15\penalty0
  (1):\penalty0 1929--1958, 2014.

\bibitem[Szegedy et~al.(2016)Szegedy, Vanhoucke, Ioffe, Shlens, and
  Wojna]{szegedy2016rethinking}
Christian Szegedy, Vincent Vanhoucke, Sergey Ioffe, Jon Shlens, and Zbigniew
  Wojna.
\newblock Rethinking the inception architecture for computer vision.
\newblock In \emph{Proceedings of the IEEE conference on computer vision and
  pattern recognition}, pp.\  2818--2826, 2016.

\bibitem[Vaswani et~al.(2017)Vaswani, Shazeer, Parmar, Uszkoreit, Jones, Gomez,
  Kaiser, and Polosukhin]{vaswani2017attention}
Ashish Vaswani, Noam Shazeer, Niki Parmar, Jakob Uszkoreit, Llion Jones,
  Aidan~N Gomez, {\L}ukasz Kaiser, and Illia Polosukhin.
\newblock Attention is all you need.
\newblock In \emph{Proceedings of the 31st International Conference on Neural
  Information Processing Systems}, pp.\  6000--6010, 2017.

\bibitem[Wan et~al.(2013)Wan, Zeiler, Zhang, Le~Cun, and
  Fergus]{wan2013regularization}
Li~Wan, Matthew Zeiler, Sixin Zhang, Yann Le~Cun, and Rob Fergus.
\newblock Regularization of neural networks using dropconnect.
\newblock In \emph{International conference on machine learning}, pp.\
  1058--1066. PMLR, 2013.

\bibitem[Wang et~al.(2018)Wang, Singh, Michael, Hill, Levy, and
  Bowman]{wang2018glue}
Alex Wang, Amanpreet Singh, Julian Michael, Felix Hill, Omer Levy, and Samuel
  Bowman.
\newblock Glue: A multi-task benchmark and analysis platform for natural
  language understanding.
\newblock In \emph{Proceedings of the 2018 EMNLP Workshop BlackboxNLP:
  Analyzing and Interpreting Neural Networks for NLP}, pp.\  353--355, 2018.

\bibitem[Wang \& Manning(2013)Wang and Manning]{wang2013fast}
Sida Wang and Christopher Manning.
\newblock Fast dropout training.
\newblock In \emph{international conference on machine learning}, pp.\
  118--126. PMLR, 2013.

\bibitem[Wang et~al.(2020)Wang, Liu, and Ma]{wang2020scale}
Yue Wang, Yuting Liu, and Zhi-Ming Ma.
\newblock The scale-invariant space for attention layer in neural network.
\newblock \emph{Neurocomputing}, 392:\penalty0 1--10, 2020.

\bibitem[Wei et~al.(2020)Wei, Kakade, and Ma]{wei2020implicit}
Colin Wei, Sham Kakade, and Tengyu Ma.
\newblock The implicit and explicit regularization effects of dropout.
\newblock In \emph{International Conference on Machine Learning}, pp.\
  10181--10192. PMLR, 2020.

\bibitem[Wen et~al.(2016)Wen, Wu, Wang, Chen, and Li]{wen2016learning}
Wei Wen, Chunpeng Wu, Yandan Wang, Yiran Chen, and Hai Li.
\newblock Learning structured sparsity in deep neural networks.
\newblock In \emph{Proceedings of the 30th International Conference on Neural
  Information Processing Systems}, pp.\  2082--2090, 2016.

\bibitem[Wu \& Gu(2015)Wu and Gu]{wu2015towards}
Haibing Wu and Xiaodong Gu.
\newblock Towards dropout training for convolutional neural networks.
\newblock \emph{Neural Networks}, 71:\penalty0 1--10, 2015.

\bibitem[Wu et~al.(2021)Wu, Zhao, and Zhang]{wu2021not}
Hongqiu Wu, Hai Zhao, and Min Zhang.
\newblock Not all attention is all you need.
\newblock \emph{arXiv preprint arXiv:2104.04692}, 2021.

\bibitem[Wu et~al.(2019)Wu, Wang, Xia, Tian, Gao, Qin, Lai, and
  Liu]{wu2019depth}
Lijun Wu, Yiren Wang, Yingce Xia, Fei Tian, Fei Gao, Tao Qin, Jianhuang Lai,
  and Tie-Yan Liu.
\newblock Depth growing for neural machine translation.
\newblock In \emph{Proceedings of the 57th Annual Meeting of the Association
  for Computational Linguistics}, pp.\  5558--5563, 2019.

\bibitem[Wu \& He(2018)Wu and He]{wu2018group}
Yuxin Wu and Kaiming He.
\newblock Group normalization.
\newblock In \emph{Proceedings of the European conference on computer vision
  (ECCV)}, pp.\  3--19, 2018.

\bibitem[Yang et~al.(2019)Yang, Dai, Yang, Carbonell, Salakhutdinov, and
  Le]{yang2019xlnet}
Zhilin Yang, Zihang Dai, Yiming Yang, Jaime Carbonell, Russ~R Salakhutdinov,
  and Quoc~V Le.
\newblock Xlnet: Generalized autoregressive pretraining for language
  understanding.
\newblock \emph{Advances in Neural Information Processing Systems},
  32:\penalty0 5753--5763, 2019.

\bibitem[Zehui et~al.(2019)Zehui, Liu, Huang, Chen, Qiu, and
  Huang]{zehui2019dropattention}
Lin Zehui, Pengfei Liu, Luyao Huang, Junkun Chen, Xipeng Qiu, and Xuanjing
  Huang.
\newblock Dropattention: A regularization method for fully-connected
  self-attention networks.
\newblock \emph{arXiv preprint arXiv:1907.11065}, 2019.

\bibitem[Zhang et~al.(2020)Zhang, Zhao, Saleh, and Liu]{zhang2020pegasus}
Jingqing Zhang, Yao Zhao, Mohammad Saleh, and Peter Liu.
\newblock Pegasus: Pre-training with extracted gap-sentences for abstractive
  summarization.
\newblock In \emph{International Conference on Machine Learning}, pp.\
  11328--11339. PMLR, 2020.

\bibitem[Zhang et~al.(2019)Zhang, Song, Gao, Chen, Bao, and Ma]{zhang2019your}
Linfeng Zhang, Jiebo Song, Anni Gao, Jingwei Chen, Chenglong Bao, and Kaisheng
  Ma.
\newblock Be your own teacher: Improve the performance of convolutional neural
  networks via self distillation.
\newblock In \emph{Proceedings of the IEEE/CVF International Conference on
  Computer Vision}, pp.\  3713--3722, 2019.

\bibitem[Zhang et~al.(2018)Zhang, Xiang, Hospedales, and Lu]{zhang2018deep}
Ying Zhang, Tao Xiang, Timothy~M Hospedales, and Huchuan Lu.
\newblock Deep mutual learning.
\newblock In \emph{Proceedings of the IEEE Conference on Computer Vision and
  Pattern Recognition}, pp.\  4320--4328, 2018.

\bibitem[Zhao et~al.(2019)Zhao, Sun, Xu, Zhang, and Luo]{zhao2019muse}
Guangxiang Zhao, Xu~Sun, Jingjing Xu, Zhiyuan Zhang, and Liangchen Luo.
\newblock Muse: Parallel multi-scale attention for sequence to sequence
  learning.
\newblock \emph{arXiv preprint arXiv:1911.09483}, 2019.

\bibitem[Zhou et~al.(2021)Zhou, Song, Chen, Zhou, Wang, Yuan, and
  Zhang]{zhou2021rethinking}
Helong Zhou, Liangchen Song, Jiajie Chen, Ye~Zhou, Guoli Wang, Junsong Yuan,
  and Qian Zhang.
\newblock Rethinking soft labels for knowledge distillation: A bias-variance
  tradeoff perspective.
\newblock \emph{International Conference on Learning Representations}, 2021.

\bibitem[Zhou et~al.(2020)Zhou, Ge, Wei, Zhou, and Xu]{zhou2020scheduled}
Wangchunshu Zhou, Tao Ge, Furu Wei, Ming Zhou, and Ke~Xu.
\newblock Scheduled drophead: A regularization method for transformer models.
\newblock In \emph{Proceedings of the 2020 Conference on Empirical Methods in
  Natural Language Processing: Findings}, pp.\  1971--1980, 2020.

\bibitem[Zhu et~al.(2019)Zhu, Xia, Wu, He, Qin, Zhou, Li, and
  Liu]{zhu2019incorporating}
Jinhua Zhu, Yingce Xia, Lijun Wu, Di~He, Tao Qin, Wengang Zhou, Houqiang Li,
  and Tieyan Liu.
\newblock Incorporating bert into neural machine translation.
\newblock In \emph{International Conference on Learning Representations}, 2019.





\bibitem[Srivastava et~al.(2014)Srivastava, Hinton,Krizhevsky,Sutskever and Salakhutdinov]{Srivastava2014DropoutAS}
Nitish Srivastava and Geoffrey E. Hinton and Alex Krizhevsky and Ilya Sutskever and Ruslan Salakhutdinov.
\newblock Dropout: a simple way to prevent neural networks from overfitting.
\newblock \emph{J. Mach. Learn. Res.}, 15:\penalty0 1929--1958, 2014.

\bibitem[Ba \& Frey(2013)Ba and Frey]{ba2013adaptive}
Lei~Jimmy Ba and Brendan Frey.
\newblock Dropout: a simple way to prevent neural networks from overfitting.
\newblock In \emph{Proceedings of the 26th International Conference on Neural
  Information Processing Systems-Volume 2}, pp.\  3084--3092, 2014.

\bibitem[Ba et~al.(2016)]{Ba2016LayerN}
Jimmy Ba and Jamie Ryan Kiros and Geoffrey E. Hinton.
\newblock Incorporating bert into neural machine translation.
\newblock In \emph{International Conference on Learning Representations}, 2019.

\bibitem[Wu et~al.(2019)Wu, Wang, Xia, Tian, Gao, Qin, Lai, and
  Liu]{wu2019depth}
Lijun Wu, Yiren Wang, Yingce Xia, Fei Tian, Fei Gao, Tao Qin, Jianhuang Lai,
  and Tie-Yan Liu.
\newblock Depth growing for neural machine translation.
\newblock In \emph{Proceedings of the 57th Annual Meeting of the Association
  for Computational Linguistics}, pp.\  5558--5563, 2019.

\bibitem[Wu \& He(2018)Wu and He]{wu2018group}
Yuxin Wu and Kaiming He.
\newblock Group normalization.
\newblock In \emph{Proceedings of the European conference on computer vision
  (ECCV)}, pp.\  3--19, 2018.

\bibitem[Vaswani et~al.(2017)Vaswani, Shazeer, Parmar, Uszkoreit, Jones, Gomez,
  Kaiser, and Polosukhin]{vaswani2017attention}
Ashish Vaswani, Noam Shazeer, Niki Parmar, Jakob Uszkoreit, Llion Jones,
  Aidan~N Gomez, {\L}ukasz Kaiser, and Illia Polosukhin.
\newblock Attention is all you need.
\newblock In \emph{Proceedings of the 31st International Conference on Neural
  Information Processing Systems}, pp.\  6000--6010, 2017.

\end{thebibliography}


\begin{thebibliography}{30}
\providecommand{\natexlab}[1]{#1}
\providecommand{\url}[1]{\texttt{#1}}
\expandafter\ifx\csname urlstyle\endcsname\relax
  \providecommand{\doi}[1]{doi: #1}\else
  \providecommand{\doi}{doi: \begingroup \urlstyle{rm}\Url}\fi

\bibitem[Ba et~al.(2016)Ba, Kiros, and Hinton]{Ba2016LayerN}
Jimmy Ba, Jamie~Ryan Kiros, and Geoffrey~E. Hinton.
\newblock Layer normalization.
\newblock \emph{ArXiv}, abs/1607.06450, 2016.

\bibitem[Cai et~al.(2019)Cai, Gao, Zhang, Wang, Chen, and
  Ooi]{Cai2019EffectiveAE}
Shaofeng Cai, Jinyang Gao, Meihui Zhang, Wei Wang, Gang Chen, and Beng~Chin
  Ooi.
\newblock Effective and efficient dropout for deep convolutional neural
  networks.
\newblock \emph{ArXiv}, abs/1904.03392, 2019.

\bibitem[Chen et~al.(2020)Chen, Gautier, and Ayd{\"o}re]{Chen2020DropClusterAS}
Liyang Chen, Philip Gautier, and Serg{\"u}l Ayd{\"o}re.
\newblock Dropcluster: A structured dropout for convolutional networks.
\newblock \emph{ArXiv}, abs/2002.02997, 2020.

\bibitem[Deng et~al.(2009)Deng, Dong, Socher, Li, Li, and
  Fei-Fei]{Deng2009ImageNetAL}
Jia Deng, Wei Dong, Richard Socher, Li-Jia Li, K.~Li, and Li~Fei-Fei.
\newblock Imagenet: A large-scale hierarchical image database.
\newblock \emph{2009 IEEE Conference on Computer Vision and Pattern
  Recognition}, pp.\  248--255, 2009.

\bibitem[Gao et~al.(2021)Gao, Yao, and Chen]{Gao2021SimCSESC}
Tianyu Gao, Xingcheng Yao, and Danqi Chen.
\newblock Simcse: Simple contrastive learning of sentence embeddings.
\newblock \emph{ArXiv}, abs/2104.08821, 2021.

\bibitem[Ghiasi et~al.(2018)Ghiasi, Lin, and Le]{Ghiasi2018DropBlockAR}
Golnaz Ghiasi, Tsung-Yi Lin, and Quoc~V. Le.
\newblock Dropblock: A regularization method for convolutional networks.
\newblock In \emph{Neural Information Processing Systems}, 2018.

\bibitem[He et~al.(2016{\natexlab{a}})He, Zhang, Ren, and Sun]{He2015DeepRL}
Kaiming He, X.~Zhang, Shaoqing Ren, and Jian Sun.
\newblock Deep residual learning for image recognition.
\newblock \emph{2016 IEEE Conference on Computer Vision and Pattern Recognition
  (CVPR)}, pp.\  770--778, 2016{\natexlab{a}}.

\bibitem[He et~al.(2016{\natexlab{b}})He, Zhang, Ren, and
  Sun]{He2016IdentityMI}
Kaiming He, X.~Zhang, Shaoqing Ren, and Jian Sun.
\newblock Identity mappings in deep residual networks.
\newblock In \emph{European Conference on Computer Vision}, 2016{\natexlab{b}}.

\bibitem[Hinton et~al.(2015)Hinton, Vinyals, and Dean]{Hinton2015DistillingTK}
Geoffrey~E. Hinton, Oriol Vinyals, and Jeffrey Dean.
\newblock Distilling the knowledge in a neural network.
\newblock \emph{ArXiv}, abs/1503.02531, 2015.

\bibitem[Hou \& Wang(2019)Hou and Wang]{Hou2019WeightedCD}
Saihui Hou and Zilei Wang.
\newblock Weighted channel dropout for regularization of deep convolutional
  neural network.
\newblock In \emph{AAAI Conference on Artificial Intelligence}, 2019.

\bibitem[Ioffe \& Szegedy(2015)Ioffe and Szegedy]{Ioffe2015BatchNA}
Sergey Ioffe and Christian Szegedy.
\newblock Batch normalization: Accelerating deep network training by reducing
  internal covariate shift.
\newblock In \emph{International Conference on Machine Learning}, 2015.

\bibitem[Khan et~al.(2017)Khan, Hayat, and Porikli]{Khan2017RegularizationOD}
Salman~Hameed Khan, Munawar Hayat, and Fatih~Murat Porikli.
\newblock Regularization of deep neural networks with spectral dropout.
\newblock \emph{Neural networks : the official journal of the International
  Neural Network Society}, 110:\penalty0 82--90, 2017.

\bibitem[Krizhevsky(2009)]{Krizhevsky2009LearningML}
Alex Krizhevsky.
\newblock Learning multiple layers of features from tiny images.
\newblock 2009.

\bibitem[Li et~al.(2022)Li, Ma, Chen, Zhang, Liu, Ma, and Yang]{Li2022ASO}
Yang~D. Li, Weizhi Ma, C.~Chen, M.~Zhang, Yiqun Liu, Shaoping Ma, and Yue Yang.
\newblock A survey on dropout methods and experimental verification in
  recommendation.
\newblock \emph{IEEE Transactions on Knowledge and Data Engineering},
  35:\penalty0 6595--6615, 2022.

\bibitem[Liang et~al.(2021)Liang, Wu, Li, Wang, Meng, Qin, Chen, Zhang, and
  Liu]{Liang2021RDropRD}
Xiaobo Liang, Lijun Wu, Juntao Li, Yue Wang, Qi~Meng, Tao Qin, Wei Chen,
  M.~Zhang, and Tie-Yan Liu.
\newblock R-drop: Regularized dropout for neural networks.
\newblock \emph{ArXiv}, abs/2106.14448, 2021.

\bibitem[Lu et~al.(2021)Lu, Xu, Du, Ishida, Zhang, and
  Sugiyama]{Lu2021LocalDropAH}
Ziqing Lu, Chang Xu, Bo~Du, Takashi Ishida, L.~Zhang, and Masashi Sugiyama.
\newblock Localdrop: A hybrid regularization for deep neural networks.
\newblock \emph{IEEE Transactions on Pattern Analysis and Machine
  Intelligence}, 44:\penalty0 3590--3601, 2021.

\bibitem[Ma et~al.(2016)Ma, Gao, Hu, Yu, Deng, and Hovy]{Ma2016DropoutWE}
Xuezhe Ma, Yingkai Gao, Zhiting Hu, Yaoliang Yu, Yuntian Deng, and Eduard~H.
  Hovy.
\newblock Dropout with expectation-linear regularization.
\newblock \emph{ArXiv}, abs/1609.08017, 2016.

\bibitem[Park \& Kwak(2016)Park and Kwak]{Park2016AnalysisOT}
Sungheon Park and Nojun Kwak.
\newblock Analysis on the dropout effect in convolutional neural networks.
\newblock In \emph{Asian Conference on Computer Vision}, 2016.

\bibitem[Pham \& Le(2021)Pham and Le]{Pham2021AutoDropoutLD}
Hieu Pham and Quoc~V. Le.
\newblock Autodropout: Learning dropout patterns to regularize deep networks.
\newblock In \emph{AAAI Conference on Artificial Intelligence}, 2021.

\bibitem[Simonyan \& Zisserman(2014)Simonyan and Zisserman]{Simonyan2014VeryDC}
Karen Simonyan and Andrew Zisserman.
\newblock Very deep convolutional networks for large-scale image recognition.
\newblock \emph{CoRR}, abs/1409.1556, 2014.

\bibitem[Srivastava et~al.(2014)Srivastava, Hinton, Krizhevsky, Sutskever, and
  Salakhutdinov]{Srivastava2014DropoutAS}
Nitish Srivastava, Geoffrey~E. Hinton, Alex Krizhevsky, Ilya Sutskever, and
  Ruslan Salakhutdinov.
\newblock Dropout: a simple way to prevent neural networks from overfitting.
\newblock \emph{J. Mach. Learn. Res.}, 15:\penalty0 1929--1958, 2014.

\bibitem[Szegedy et~al.(2016)Szegedy, Vanhoucke, Ioffe, Shlens, and
  Wojna]{Szegedy2015RethinkingTI}
Christian Szegedy, Vincent Vanhoucke, Sergey Ioffe, Jon Shlens, and Zbigniew
  Wojna.
\newblock Rethinking the inception architecture for computer vision.
\newblock In \emph{Proceedings of the IEEE conference on computer vision and
  pattern recognition}, pp.\  2818--2826, 2016.

\bibitem[Tompson et~al.(2015)Tompson, Goroshin, Jain, LeCun, and
  Bregler]{Tompson2014EfficientOL}
Jonathan Tompson, Ross Goroshin, Arjun Jain, Yann LeCun, and Christoph Bregler.
\newblock Efficient object localization using convolutional networks.
\newblock \emph{2015 IEEE Conference on Computer Vision and Pattern Recognition
  (CVPR)}, pp.\  648--656, 2015.

\bibitem[Wu \& Gu(2015)Wu and Gu]{Wu2015TowardsDT}
Haibing Wu and Xiaodong Gu.
\newblock Towards dropout training for convolutional neural networks.
\newblock \emph{Neural networks : the official journal of the International
  Neural Network Society}, 71:\penalty0 1--10, 2015.

\bibitem[Wu \& He(2018)Wu and He]{Wu2018GroupN}
Yuxin Wu and Kaiming He.
\newblock Group normalization.
\newblock In \emph{Proceedings of the European conference on computer vision
  (ECCV)}, pp.\  3--19, 2018.

\bibitem[Zeng et~al.(2021)Zeng, Dai, Chen, Xia, and
  Lu]{Zeng2021CorrelationbasedSD}
Yuyuan Zeng, Tao Dai, Bin Chen, Shutao Xia, and Jian Lu.
\newblock Correlation-based structural dropout for convolutional neural
  networks.
\newblock \emph{Pattern Recognit.}, 120:\penalty0 108117, 2021.

\bibitem[Zhang et~al.(2019)Zhang, Song, Gao, Chen, Bao, and Ma]{Zhang2019BeYO}
Linfeng Zhang, Jiebo Song, Anni Gao, Jingwei Chen, Chenglong Bao, and Kaisheng
  Ma.
\newblock Be your own teacher: Improve the performance of convolutional neural
  networks via self distillation.
\newblock In \emph{Proceedings of the IEEE/CVF International Conference on
  Computer Vision}, pp.\  3713--3722, 2019.

\bibitem[Zhang et~al.(2018)Zhang, Xiang, Hospedales, and Lu]{zhang2018deep}
Ying Zhang, Tao Xiang, Timothy~M Hospedales, and Huchuan Lu.
\newblock Deep mutual learning.
\newblock In \emph{Proceedings of the IEEE Conference on Computer Vision and
  Pattern Recognition}, pp.\  4320--4328, 2018.

\bibitem[Zolna et~al.(2017)Zolna, Arpit, Suhubdy, and
  Bengio]{Zolna2017FraternalD}
Konrad Zolna, Devansh Arpit, Dendi Suhubdy, and Yoshua Bengio.
\newblock Fraternal dropout.
\newblock \emph{ArXiv}, abs/1711.00066, 2017.

\bibitem[Zoph et~al.(2017)Zoph, Vasudevan, Shlens, and Le]{Zoph2017LearningTA}
Barret Zoph, Vijay Vasudevan, Jonathon Shlens, and Quoc~V. Le.
\newblock Learning transferable architectures for scalable image recognition.
\newblock \emph{2018 IEEE/CVF Conference on Computer Vision and Pattern
  Recognition}, pp.\  8697--8710, 2017.

\end{thebibliography}

\clearpage


\end{document}